\documentclass{arxiv_bmvc2k}
\usepackage{soul}      
\usepackage{amssymb}    
\usepackage{bbm}       
\usepackage{relsize}    
\usepackage{bbding}   
\usepackage{multirow}

\title{Grounded Situation Recognition\\with Transformers}

\addauthor{Junhyeong Cho*}{junhyeong99@postech.ac.kr}{1}
\addauthor{Youngseok Yoon*}{yys8646@postech.ac.kr}{1}
\addauthor{Hyeonjun Lee*}{hyeonjun1882@postech.ac.kr}{2}
\addauthormoreinst{Suha Kwak}{suha.kwak@postech.ac.kr}{1}{2}

\addinstitution{
    Department of CSE\\
    POSTECH\\
    Pohang, Republic of Korea
}
\addinstitution{
    Graduate School of AI\\
    POSTECH\\
    Pohang, Republic of Korea
}

\runninghead{Cho et al.}{Grounded Situation Recognition with Transformers}

\def\ie{\emph{i.e}\bmvaOneDot}
\def\eg{\emph{e.g}\bmvaOneDot}

\def\etal{\emph{et al}\bmvaOneDot}

\definecolor{brown}{rgb}{0.65, 0.16, 0.16}
\definecolor{purp}{rgb}{0.65, 0.16, 0.65}
\definecolor{orange}{rgb}{1.0, 0.5, 0.0}
\definecolor{black}{rgb}{0.0, 0.0, 0.0}

\begin{document}

\maketitle
\begin{abstract}
    Grounded Situation Recognition (GSR) is the task that not only classifies a
        salient action (\textit{verb}), but also predicts entities (\textit{nouns}) associated with semantic roles and
        their locations in the given image. 
    Inspired by the remarkable success of Transformers in vision tasks, 
        we propose a GSR model based on a Transformer encoder-decoder architecture.
    The attention mechanism of our model enables accurate verb classification by capturing high-level semantic feature of an image effectively, and allows the model to flexibly deal with the complicated and image-dependent relations between entities for improved noun classification and localization.
    Our model is the first Transformer architecture for GSR, and achieves the state of the art in every evaluation metric on the SWiG benchmark. Our code is available at \url{https://github.com/jhcho99/gsrtr}.
    
\end{abstract}
\begin{figure}[h!]
\centering
\includegraphics[height=3.61cm]{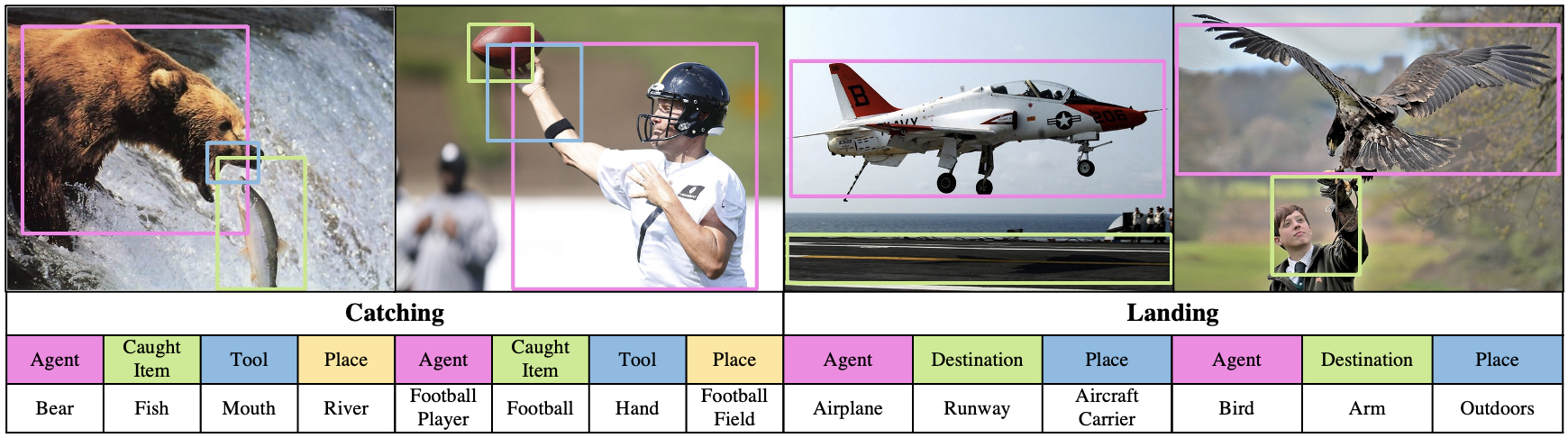}
\caption{Predictions of our model on the SWiG dataset.
}
\label{fig:teaser}
\end{figure}

\section{Introduction}
\label{sec:intro}

\begin{figure}[!t]
\centering
   \includegraphics[width=\textwidth]{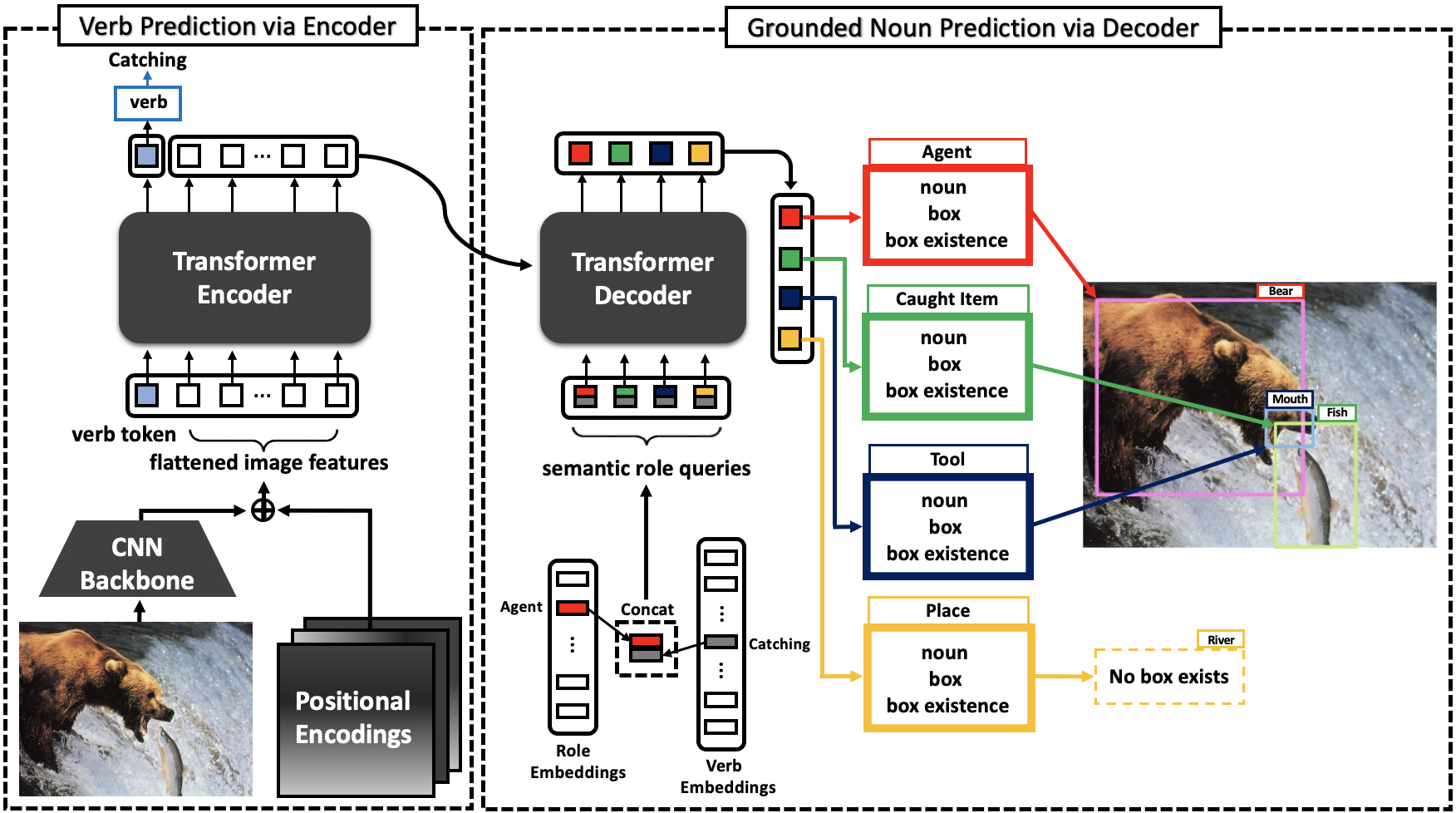}
\caption{
The overall architecture of our model (GSRTR). It mainly consists of two components: 
Transformer Encoder for verb prediction, and Transformer Decoder for grounded noun prediction.
Diagram is best viewed in colored version.
}
\label{fig:overall}
\end{figure}

Deep learning models have achieved or even surpassed human-level performance on basic vision tasks such as classification of objects~\cite{resnet,pham2021meta}, actions~\cite{zhao2017single,safaei2019still}, and places~\cite{zhou2014learning,gong2014multi,liu2019novel}.
However, it still remains challenging and less explored 
to expand such models for detailed and comprehensive understanding of natural scenes, \eg, recognizing what happens and who are involved with which roles.
Image captioning~\cite{vinyals2015show,you2016image,huang2019attention} and scene graph generation~\cite{xu2017scene,yang2018graph,liu2021fully} have been studied in this context. 
These tasks aim at reasoning about image contents in detail and describing them through natural language captions or relation graphs of objects.
However, 
quality evaluation of natural language captions is not straightforward, 
and scene graphs are limited in terms of expressive power as they represent an action only by a triplet of subject, predicate, and object.

Grounded Situation Recognition (GSR)~\cite{pratt2020grounded} is a comprehensive scene understanding task that resolves the above limitations. 
It originates from Situation Recognition (SR)~\cite{yatskar2016situation}, the task of predicting a salient action, entities taking part of the action, and their roles altogether given an image. 
In SR, an action and entities are called \emph{verb} and \emph{nouns}, respectively, and the set of semantic \emph{roles} of the entities in an action is termed \emph{frame}; a frame is defined for each verb as prior knowledge by FrameNet~\cite{fillmore2003background}, a lexical database of English.
Then SR is typically done by predicting a verb then assigning a noun to each role given by the frame of the verb. 
GSR has been introduced to further address localization (\ie, bounding box estimation) of the nouns in the image, which is missing in SR.
It is thus more challenging yet enables more detailed scene understanding in comparison with SR. 

The major challenge in GSR is two-fold.
The first is the difficulty of verb prediction. 
This is caused by the fact that a verb is a high-level concept embodied by multiple entities; as illustrated in Fig.~\ref{fig:teaser}, images of the same verb often vary significantly due to different entities interacting in different ways. 
The second is the difficulty of modeling complicated relations between entities.
Since an action (\ie, verb) is performed by multiple entities (\ie, nouns) related to each other, individual noun recognition per role is definitely suboptimal; relations between nouns have to be considered for improved noun prediction and localization.
However, modeling such relations is challenging since they are latent and depending on an input image.

Inspired by the recent success of Transformers~\cite{vaswani2017attention, dosovitskiy2021an, carion2020end}, we present in this paper a new model, dubbed GSRTR, that addresses the aforementioned challenges through the attention mechanism.
As illustrated in Fig.~\ref{fig:overall}, it has an encoder-decoder architecture based on Transformer.
The encoder takes as input a verb token and image features from a CNN backbone.
The token then goes through self-attention blocks in the encoder and is finally processed by a verb classifier on top.
Thanks to the self-attention with the image features, the encoder can capture rich and high-level semantic information for accurate verb prediction.
Meanwhile, the decoder predicts a grounded noun per role, where target roles are determined by the frame of the target verb.
It thus takes as input \emph{semantic role queries} of target roles as well as image features given by the encoder;
a semantic role query is obtained by a concatenation of two embedding vectors, one for its role and the other for a verb, which are learnable parameters dedicated to the role and verb, respectively.
Each semantic role query is converted to a feature vector through attention blocks, then used to predict a noun class, a box coordinate and a box existence probability of its role.
The attention blocks in our decoder allow to capture complicated and image-dependent relations among roles effectively and flexibly.

\noindent\textbf{Contributions:} Our GSRTR is the first Transformer architecture dedicated to GSR.
Furthermore, its encoder-decoder architecture is carefully designed to address major challenges of the task. 
The efficacy of GSRTR is validated on the SWiG dataset~\cite{pratt2020grounded}, the standard benchmark for GSR, where it clearly outperforms existing models~\cite{pratt2020grounded} in every evaluation metric.
We also provide in-depth analysis on behaviors of GSRTR, which demonstrates that it has the capability of drawing attentions on local areas relevant to verb and grounded nouns.

\section{Related~Work}

\textbf{Situation Recognition:}
Situation Recognition (SR) is the task of predicting a salient action (\textit{verb}) and entities (\textit{nouns}) taking part of the action.
Yatskar~\etal~\cite{yatskar2016situation} present the \textit{imSitu} dataset as benchmark of Situation Recognition and propose Conditional Random Field (CRF) model.
Their following work~\cite{yatskar2017commonly} figures out that sparsity of training examples compared to large output space could be problematic, 
and alleviates it through tensor-composition function.
Since then, there have been attempts to model the relations among semantic roles.
Inspired by image captioning task, Mallya and Lazebnik~\cite{mallya2017recurrent} adopt a Recurrent Neural Network (RNN) architecture to model the relations in the predefined order.
Li \etal~\cite{li2017situation} use a Gated Graph Neural Network (GGNN)~\cite{li2016gated} to capture relations among roles, and 
Suhail and Sigal~\cite{suhail2019mixture} propose a modified GGNN to learn context-aware relations among roles depending on the content of the image. 
Cooray \etal~\cite{cooray2020attention} formulate the relation modeling as an interdependent query based visual reasoning problem.

\noindent \textbf{Grounded Situation Recognition:}
Recently, Grounded Situation Recognition (GSR) has been introduced by Pratt 
    \etal~\cite{pratt2020grounded} to further address localization of entities, which is missing in SR. 
They propose the Situation With Groundings (SWiG) dataset that provides bounding box annotations in addition to the \textit{imSitu} dataset.
They also propose Joint Situation Localizer (JSL) model which consists of a verb classifier and a RNN based object detector.
The object detector sequentially produces noun and its bounding box prediction via the predefined role order.
Compared with JSL, our GSRTR can flexibly capture the relations among the semantic roles rather than the predefined order.
Furthermore, the verb prediction process in our model can capture long-range interactions of semantic concepts via a Transformer encoder.

\noindent\textbf{Transformer in Vision Tasks:} Dosovitskiy \etal~\cite{dosovitskiy2021an} propose a standard Transformer encoder architecture~\cite{vaswani2017attention} for image classification task.
This model, called ViT, takes image patches flattened, linearly transformed, and combined with positional encodings as input with a classification token.
On the other hand, the encoder of GSRTR takes image features from a CNN backbone as input, and is combined with a decoder for grounded noun prediction.
Carion \etal~\cite{carion2020end} view object detection as a direct set prediction and bipartite matching problem, and propose a Transformer encoder-decoder architecture for object detection accordingly.
Their model, called DETR, introduces learnable embeddings called \textit{object queries} as inputs of the decoder, each of which is in charge of a certain image region and a set of bounding box candidates.
Instead of the object queries, GSRTR uses \emph{semantic role queries}, each of which focuses on entities taking part of a specified action with a specific role.

\noindent\textbf{Similar follow-ups to DETR:}
 There have been attempts, including our GSRTR, to apply DETR to other domains such as video instance segmentation \cite{wang2021end}, video action recognition \cite{zhang2021temporal} and human-object-interaction detection \cite{zou2021end}.
Their models use latent queries for a Transformer decoder in the similar way, but GSRTR has notable differences.
While their models employ a fixed number of latent queries in the decoder, GSRTR constructs a variable number of queries depending on a given image.
Also, to the best of our knowledge, GSRTR is the first attempt to explicitly leverage the output of a Transformer encoder for building queries used in a Transformer decoder; semantic role queries use the verb embedding corresponding to the predicted verb from the encoder output at inference time.

\section{Proposed Method}
Inspired by ViT~\cite{dosovitskiy2021an} and DETR~\cite{carion2020end}, we propose a novel model called Grounded Situation Recognition TRansformer (GSRTR) to address the challenging GSR task;
the architecture of GSRTR is illustrated in Fig.~\ref{fig:overall}.
This section first provides a formal definition of GSR, then describes details of our model architecture, training and inference procedures.

\subsection{Task Definition}
\label{sec:def}
Let $\mathcal V$, $\mathcal R$, and $\mathcal N$ denote the sets of verbs, roles, and nouns defined in the task, respectively. 
For each verb $v \in \mathcal V $, a set of semantic roles, denoted by $\mathcal R_v \subset \mathcal R$, is predefined as its frame by FrameNet~\cite{fillmore2003background}.
For example, the frame of a verb $Catching$ is a set of semantic roles $\mathcal R_{Catching} =$ \{$Agent,$ $Caught$ $Item,$ $Tool,$ $Place$\} $\subset \mathcal R$.
Also, a pair of a noun $n\in\mathcal{N}$ and its bounding box $\mathbf{b} \in \mathbb{R}^4$ is called a \emph{grounded noun}.
The goal of GSR is to predict a verb $v$ of an input image and assign a grounded noun to each role in $\mathcal{R}_v$.
Formally speaking, a prediction of GSR is in the form of
$S = (v, \mathcal{F}_v)$, where 
$\mathcal F_v=\{ \left( r, n_r, \mathbf{b}_r \right) \vert\;
        n_r \in \mathcal N \cup \left\{ \emptyset_n \right\},\;
        \mathbf{b}_r \in \mathbb{R}^4 \cup \left\{ \emptyset_b \right\} \;
        \mathrm{for} \; r\in\mathcal R_v\}$;
$\emptyset_n$ and $\emptyset_b$ mean \emph{unknown} and \emph{not grounded}, respectively.
For example, the prediction for the leftmost image in Fig.~\ref{fig:teaser} is given by
    $S = \big(Catching,$
    $\big\{(Agent,$ $Bear,$ $\textcolor{red}{\Box}),$
    $(Caught$ $Item,$ $Fish,$ $\textcolor{green}{\Box}),$
    $(Tool,$ $Mouth,$ $\textcolor{blue}{\Box}),$
    $(Place,$ $River,$ $\emptyset_b)\big\}\big)$.

\subsection{Encoder for Verb Prediction}
\label{sec:verb}
A CNN backbone first processes an input image to extract its feature map $X_{img} \in \mathbb R^{c \times h \times w}$, where $c$ is the number of channels and $h \times w$ is the resolution of $X_{img}$.
Then $X_{img}$ is fed to a $1\times1$ convolution layer for reducing the channel size to $d$, and flattened, leading to flattened images features $F_{img} \in \mathbb R^{d \times hw}$.
Like the classification token used in ViT~\cite{dosovitskiy2021an}, we append a learnable verb embedding $\mathbf{f}_{v} \in \mathbb R^{d}$ to $F_{img}$, forming an input of the encoder $F \in \mathbb R^{d \times (1+hw)}$.

The encoder is a stack of six layers, each of which consists of a Multi-Head Self-Attention (MHSA) block and a Feed Forward Network (FFN) block.
Also, we apply Pre-Layer Normalization (Pre-LN)~\cite{xiong2020layer} before the MHSA and FFN blocks. Positional encodings are added to the input of each encoder layer.
Please refer to the supplementary material for more details of the encoder.

The output of the encoder, denoted by $E \in \mathbb R^{d \times (1+hw)}$, is split into a verb feature $\mathbf{e}_{v} \in \mathbb R^{d}$ and $hw$ image features $E_{img} \in \mathbb R^{d \times hw}$.
The former is fed to the verb classifier, which in turn produces a logit vector $\mathbf {z}_{v} \in \mathbb R^{\vert \mathcal V \vert}$ as a result of verb classification. 
On the other hand, the latter will be used as observations for the decoder.
Note that by exploiting the attention mechanism through the encoder layers, the verb token can effectively aggregate relevant semantic features of an image for accurate verb classification.

\subsection{Decoder for Grounded Noun Prediction}
\label{sec:noun}
In addition to the image features $E_{img}$ given by the encoder, the decoder takes as input semantic role queries to predict corresponding nouns and their bounding boxes, inspired by the object queries in DETR~\cite{carion2020end}.
To be specific, a semantic role query $\mathbf{w}_{(v,r)}\in\mathbb{R}^d$ is obtained by a concatenation of 
    a verb embedding vector $\mathbf{w}_{v} \in \mathbb R^{d_{v}}$
    and
    a role embedding vector $\mathbf{w}_{r} \in \mathbb R^{d_{r}}$
    ($d = d_{v} + d_{r}$),
    both of which are learnable parameters;
$v$ is the ground-truth verb at training time and the predicted verb at inference time, while $r\in\mathcal R_v$.
The number of semantic role queries fed to the decoder is thus $|\mathcal R_v|$.

The decoder is a stack of six layers, each of which consists of a MHSA block, a Multi-Head Attention (MHA) block, and a FFN block; Pre-LN is applied before each of the blocks.
The first decoder layer input is set to zero.
In each decoder layer, each semantic role query $\mathbf{w}_{(v,r)}$ is added to each key and query of the MHSA block and added to each query of the MHA block.
The image features $E_{img}$ serve as keys and values in the MHA block of each decoder layer.
Through the MHSA block in each decoder layer, semantic role queries flexibly capture the role relations (Fig.~\ref{fig:roles}).
From the MHA block in each decoder layer, each semantic role query attends to image features considering image-dependent relations (Fig.~\ref{fig:role_img}).

Through the decoder, each semantic role query $\mathbf{w}_{(v, r)}$ is converted to an output feature. 
The output feature of each role $r\in\mathcal R_v$ is in turn fed to three branches:
One for noun classification, another for bounding box regression, and the other for predicting existence of its bounding box.
The noun classifier produces a noun logit vector $\mathbf{z}_{n_{r}} \in \mathbb R^{\vert \mathcal N \cup \{\emptyset_n\}\vert}$.
The bounding box regressor predicts $\mathbf{\hat b}_{r}^\prime = (\hat c_x, \hat c_y, \hat w, \hat h) \in [0,1]^{4}$, indicating the normalized center coordinate, height, and width of a box relative to the image size.
This predicted box coordinate is transformed into top-left and bottom-right coordinate representation $\mathbf{\hat b}_{r} = (\hat x_1, \hat y_1, \hat x_2, \hat y_2) \in \mathbb R^4$.
Finally, the box existence predictor produces a box existence probability $p_{{b}_{r}} \in [0,1]$.
Please refer to the supplementary material for more details of the decoder.

\subsection{Training and Inference}
\label{sec:procedure}
The total loss for training GSRTR is a linear combination of five losses: A verb classification loss, a noun classification loss, a bounding box existence loss, 
a $L_1$ box regression loss, and a Generalized IoU (GIoU)~\cite{rezatofighi2019generalized} box regression loss.
The verb classification loss $\mathcal L_v$ is the cross entropy between the verb prediction probability $\mathbf{p}_v=\mathrm{Softmax}(\mathbf{z}_{v})$ and the ground-truth verb distribution.
The noun classification loss $\mathcal L_n$ is formulated as the average of individual noun classification losses over the semantic roles, and is given by
\begin{align}
    \label{eq:loss_n}
    \mathcal L_{n} &= \frac{1}{\vert \mathcal R_v \vert} \sum_{r\in \mathcal R_v} \mathrm{CrossEntropy}(\mathbf{p}_{n_{r}}, \mathbf{t}_{n_{r}}),
\end{align}
where $\mathbf{p}_{n_{r}}$ denotes the noun prediction probability for each role $r$ and $\mathbf{t}_{n_{r}}$ indicates the ground-truth noun distribution for each role $r$.
The bounding box existence loss $\mathcal L_{exist}$ is the average of individual bounding box existence loss over the semantic roles, and is given by
\begin{align}
    \label{eq:loss_exist}
    \mathcal L_{exist} &= \frac{1}{\vert \mathcal R_v \vert} \sum_{r\in\mathcal R_v} \mathrm{CrossEntropy} (p_{{b}_{r}}, t_{{b}_{r}}),
\end{align}
where $p_{{b}_{r}}$ denotes the bounding box existence probability for each role $r$ and 
$t_{{b}_{r}} \in \{0,1\}$ specifies the existence of the ground-truth bounding box for each role $r$ (\emph{i.e.}, $t_{{b}_r}=1$ when $\mathbf{b}_r \neq \emptyset_b$).
The $L_1$ box regression loss $\mathcal L_{L_1}$ is defined as the average of individual $L_1$ distances between predicted and ground-truth bounding boxes  over semantic roles for which ground-truth bounding boxes exist, and are given by
\begin{align}
    \label{eq:loss_l1}
    \mathcal L_{L_1} &= \frac{1}{\vert \mathcal{\tilde R}_v \vert} \mathlarger{\sum}_{r \in \mathcal{\tilde R}_v} \Vert \mathbf{\hat b}_{r}^\prime - \mathbf{b}_r^\prime \Vert_1,
\end{align}
where $\mathcal{\tilde R}_v = \{r \;\vert\; r\in\mathcal R_v \; \mathrm{and} \; \mathbf{b}_r \neq \emptyset_b \}$ is the set of roles associated with bounding boxes.
Finally, the GIoU box regression loss $\mathcal L_{GIoU}$~\cite{rezatofighi2019generalized} is formulated as the average of individual GIoU losses over roles for which ground-truth bounding boxes exist, and are given by
\begin{align}
    \label{eq:loss_giou}
    \mathcal L_{GIoU} &= \frac{1}{\vert \mathcal{\tilde R}_v \vert} \mathlarger{\sum}_{r\in \mathcal{\tilde R}_v}
    \left( 1 - \left( 
        \frac{\vert \mathbf{b}_r \cap \mathbf{\hat b}_{r} \vert}{\vert \mathbf{b}_r \cup \mathbf{\hat b}_r \vert}
        - \frac{\vert C(\mathbf{b}_r, \mathbf{\hat b}_{r}) \setminus \mathbf{b}_r \cup \mathbf{\hat b}_{r} \vert}{\vert C(\mathbf{b}_r, \mathbf{\hat b}_{r})\vert}
        \right)
    \right),
\end{align}
where $C(\mathbf{\hat b}_r, \mathbf{b}_r)$ denotes the smallest box enclosing predicted box $\mathbf{\hat b}_r$ and ground-truth box $\mathbf{b}_r$ for each role $r$.
GIoU loss is a scale-invariant loss and it compensates for scale-variant $L_1$ loss.
The total loss $\mathcal L_{total}$ is formulated as $\mathcal L_{total} = \lambda_v \mathcal L_{v} + \lambda_n \mathcal L_{n} + \lambda_{exist} \mathcal L_{exist} + \lambda_{L_1} \mathcal L_{L_1} + \lambda_{GIoU} \mathcal L_{GIoU}$, where $\lambda_v,\lambda_n,\lambda_{exist},\lambda_{L_1},\lambda_{GIoU} > 0$ are hyperparameters.

At inference time, our method predicts a verb $\hat{v} = \arg\max_{v} \mathbf{p}_v$ then constructs corresponding semantic role queries $\mathbf{w}_{(\hat{v},r)}$ for all $r \in \mathcal R_{\hat v}$.
Each $\mathbf{w}_{(\hat v, r)}$ is used by the decoder to produce corresponding output noun logit $\mathbf{z}_{n_{r}}$, bounding box $\mathbf{\hat b}_{r}^\prime$ and bounding box existence probability $p_{{b}_{r}}$.
Note that if $p_{{b}_{r}}<0.5$, the predicted bounding box $\mathbf{\hat b}_{r}^\prime$ is ignored.

\section{Experiments}

\subsection{Dataset and Metrics}
\label{exp:data}
SWiG~\cite{pratt2020grounded} dataset is composed of 75k, 25k and 25k images for the train, development and test set respectively.
There are $\vert \mathcal V \vert = 504$ verbs, $\vert \mathcal R \vert = 190$ roles, and $1 \le \vert \mathcal R_v \vert \le 6$ semantic roles per verb.
We use about $10k$ nouns, the number of noun classes in the train set.
The annotation for each image consists of a verb, a bounding box for each semantic role, and three nouns (from three annotators) for each semantic role.

The predicted verb and grounded nouns are measured by five metrics: \textit{verb}, \textit{value}, \textit{value-all}, \textit{grounded-value}, and \textit{grounded-value-all}.
The \textit{verb} metric denotes a verb prediction accuracy.
The \textit{value} metric denotes a noun prediction accuracy from its semantic role.
The \textit{value-all} metric denotes that all nouns corresponding to semantic roles are correctly predicted.
The \textit{grounded-value} metric denotes a grounded noun prediction accuracy for its semantic role.
Note that the grounded noun prediction is considered correct if it correctly predicts noun and bounding box.
The bounding box prediction is considered correct if it correctly predicts bounding box existence and the predicted box has an Intersection-over-Union (IoU) value of at least 0.5 with the ground-truth box.
The \textit{grounded-value-all} metric denotes that all grounded nouns corresponding to semantic roles are correctly predicted. 
The requirements for each metric are summarized in Table~\ref{table:metric}.
Because the number of roles per verb is different and the number of images per verb could be different, all above metrics are calculated for each verb and then averaged over them. 

Since these metrics depend heavily on the verb accuracy, the metrics are reported in 3 settings: \textbf{top-1 predicted verb}, \textbf{top-5 predicted verbs} and \textbf{ground-truth verb}.
In \textbf{top-1 predicted verb} setting, five metrics are reported: a top-1 predicted verb accuracy, two noun metrics and two grounded noun metrics.
If the top-1 predicted verb is incorrect, the noun and grounded noun metrics are considered incorrect.
In \textbf{top-5 predicted verbs} setting, five metrics are reported: a top-5 predicted verbs accuracy, two noun metrics and two grounded noun metrics.
If the ground-truth verb is not included in the top-5 predicted verbs, the noun and grounded noun metrics are
considered incorrect, too.
In \textbf{ground-truth verb} setting, four metrics are reported: two noun metrics and two grounded noun metrics.
From the ground-truth verb assumed to be known, noun and grounded noun predictions are taken from the model by conditioning on the ground-truth verb.

\begin{table}[t]
\centering
\caption{Requirements for each metric.} 
\centering 
\label{table:metric}
\resizebox{\textwidth}{!}{
\begin{tabular}{c|c|c|c|c|c} 
    \hline 
    \multicolumn{1}{c|}{} & \multicolumn{5}{c}{requirement} \\
    \hline 
    & \multirow{2}{*}{correct verb}
    & correct noun
    & correct nouns
    & correct bounding box
    & correct bounding boxes
    \\
    metric &
    &
    for a semantic role & 
    for all semantic roles &
    for a semantic role &
    for all semantic roles
    \\
\hline 
\hline
\textit{verb} & \Checkmark &  &  & \\ 
\hline
\textit{value} & \Checkmark & \Checkmark &  & \\
\hline
\textit{value-all} & \Checkmark & \Checkmark & \Checkmark & \\
\hline
\textit{grounded-value} & \Checkmark & \Checkmark &  &\Checkmark \\
\hline
\textit{grounded-value-all} & \Checkmark & \Checkmark & \Checkmark & \Checkmark & \Checkmark \\ 
\hline 
\end{tabular}}
\end{table}

\subsection{Implementation Details}
\label{exp:detail}

Following previous work~\cite{pratt2020grounded}, we use ImageNet-pretrained ResNet-50 backbone~\cite{resnet} except Feature Pyramid Network (FPN) ~\cite{lin2017_fpn}.
The ResNet-50 backbone produces the image features $X_{img} \in \mathbb R ^ {c \times h \times w}$ from the input image where $c=2048$.
The hidden dimensions of each semantic role query, verb token and image feature are $512$ ($d=512$).
The verb embedding dimension and role embedding dimension are $256$ ($d_v=d_r=256$).
We use learnable 2D embeddings for the positional encodings.
The number of heads for all MHSA and MHA blocks is $8$.
We use 2 fully connected layers with ReLU activation function for the four followings: the FFN blocks in the encoder and decoder, the verb classifier, the noun classifier, and the bounding box existence predictor.
The size of hidden dimensions are $2048$, $2d$, $2d$, and $2d$, respectively.
The dropout rates are $0.15$, $0.3$, $0.3$, and $0.2$, respectively.
The bounding box regressor is 3 fully connected layers with ReLU activation function and $2d$ hidden dimensions, using $0.2$ dropout rate.
The label smoothing regularization~\cite{szegedy2016rethinking} is used for the target verb and noun labels with 
label smoothing factor $0.3$ and $0.2$, respectively.
We use AdamW~\cite{loshchilov2018decoupled} optimizer with the learning rate $10^{-4}$ ($10^{-5}$ for the backbone),
weight decay $10^{-4}$, $\beta_{1}=0.9$ and $\beta_{2}=0.999$.
We set the max gradient clipping value to $0.1$ and train the BatchNorm layers in the backbone.
The training epoch is 40 with batch size 16 per GPU on four 12GB TITAN Xp GPUs, which takes about 20 hours.
The loss coefficients are $\lambda_v=\lambda_n=1$ and $\lambda_{exist}=\lambda_{L_1}=\lambda_{GIoU}=5$.

\noindent\textbf{Data Augmentation:} Random Color Jittering, Random Gray Scaling, Random Scaling and Random Horizontal Flipping are used.
The hue, saturate and brightness scale in random color jittering set to $0.1$.
The scale of random gray scaling sets to $0.3$.
The scales of random scaling set to $0.5$, $0.75$ and $1.0$.
The probability of random horizontal flipping sets to $0.5$.

\noindent\textbf{Final Noun Loss:} In SWiG, three noun annotations exist per role. For each noun annotation, we calculate the loss~(Eq. \ref{eq:loss_n}). The final noun loss is the summation of the three noun losses. 

\noindent \textbf{Batch Training:} 
The number of semantic roles ranges from 1 to 6 depending on the frame of a verb.
In GSRTR, the semantic role queries are constructed as much as the number of semantic roles.
To ensure batch training, zero padding is used for each output of grounded noun prediction branches. 
We ignore the padded outputs in the loss computation.

\subsection{Experiment Results}
\textbf{Quantitative Comparison with Previous Work:}
Table~\ref{table:result} quantitatively compares our model with previous work on the \emph{dev} and \emph{test} splits of SWiG dataset.
In all evaluation metrics, GSRTR achieves the state-of-the-art accuracy.
In the \emph{dev} set, compared with JSL, GSRTR achieves the top-1 predicted verb and top-5 predicted verbs accuracies of 41.06\% (+1.46\%p) and 69.46\% (+1.75\%p), respectively.
In ground-truth verb setting, GSRTR achieves the value and grounded-value accuracies 74.27\% (+0.74\%p) and 58.33\% (+0.83\%p), respectively.
Note that previous work uses two ResNet-50 backbones and FPN, while our GSRTR only uses a single ResNet-50 backbone without FPN. 
Existing models in~\cite{pratt2020grounded} have about 108 million parameters, but our GSRTR only has about 83 million parameters.
Although GSRTR has less backbone capacity and less parameters, it achieves the state-of-the-art accuracy in every evaluation metric. In addition, the reason for the small improvement by GSRTR in terms of grounded-value metrics is that these metrics require correct predictions of verb, noun and bounding box as shown in Table~\ref{table:metric}.

Existing models in~\cite{pratt2020grounded} are trained separately in terms of verb prediction part and grounded noun prediction part, while our GSRTR is trained in an end-to-end manner.
For this reason, it is difficult to fairly compare the training time of ours with existing models.
However, we can reasonably guess that GSRTR takes less training time than others.
GSRTR takes about 20 hours with four 12GB TITAN Xp GPUs for whole training, but other models take about 20 hours with four 24GB TITAN RTX GPUs only for training of grounded noun prediction part.
For the comparison of inference time, we compare GSRTR with JSL which was the previous state-of-the-art.
We evaluate the models on the \emph{test} set in the same environment with one 2080Ti GPU.
GSRTR takes 21.69 ms (46.10 FPS) and JSL takes 80.00 ms (12.50 FPS) on the average of 10 trials.
\label{exp:result}

\noindent
\textbf{Effect of Verb Embedding Concatenation:}
We also quantitatively show the effect of verb embedding concatenation in the semantic role query.
If we do not concatenate the verb embedding (\emph{i.e.}, $d_v=0$ and $d_r=d$), the accuracies in the ground-truth verb setting decrease by around $1.3 \sim 2.3$\%p  (GSRTR w/o VE in Table~\ref{table:result}).
It demonstrates that the verb embedding concatenation is helpful for grounded noun prediction.

\begin{table}[!t]
    \centering
    \caption{
        Quantitative evaluation on the SWiG dataset.
    }
    \label{table:result}
    \resizebox{\textwidth}{!}{
        \begin{tabular}{l|l|ccccc|ccccc|cccc}
        \hline
        \multicolumn{2}{c|}{}
            & \multicolumn{5}{c|}{top-1 predicted verb}
            & \multicolumn{5}{c|}{top-5 predicted verbs}
            & \multicolumn{4}{c}{ground-truth verb}  
        \\
        \hline
            &  
            &       &       &       & grnd & grnd
            &       &       &       & grnd & grnd
            &       &       & grnd & grnd  
        \\
        set & model 
            & verb & value & value-all & value & value-all
            & verb & value & value-all & value & value-all
            & value & value-all & value & value-all
        \\
        \hline
        \hline
        \multirow{4}{*}{dev} & ISL \cite{pratt2020grounded}
            & 38.83 & 30.47 & 18.23 & 22.47 & 7.64
            & 65.74 & 50.29 & 28.59 & 36.90 & 11.66
            & 72.77 & 37.49 & 52.92 & 15.00 
        \\
            & JSL \cite{pratt2020grounded}
            & 39.60 & 31.18 & 18.85 & 25.03 & 10.16
            & 67.71 & 52.06 & 29.73 & 41.25 & 15.07
            & 73.53 & 38.32 & 57.50 & 19.29
        \\
        \cline{2-16}
            & GSRTR w/o VE (Ours)
            & 40.81 & 32.05 & 19.31 & 25.64 & 10.31
            & 69.33 & 53.09 & 29.78 & 42.01 & 15.36
            & 72.55 & 37.07 & 57.00 & 18.93
        \\
            & GSRTR (Ours)
            & \textbf{41.06} & \textbf{32.52} & \textbf{19.63} & \textbf{26.04} & \textbf{10.44}
            & \textbf{69.46} & \textbf{53.69} & \textbf{30.66} & \textbf{42.61} & \textbf{15.98}
            & \textbf{74.27} & \textbf{39.24} & \textbf{58.33} & \textbf{20.19} 
        \\
        \hline
        \hline
        \multirow{4}{*}{test} & ISL \cite{pratt2020grounded}
            & 39.36 & 30.09 & 18.62 & 22.73 & 7.72
            & 65.51 & 50.16 & 28.47 & 36.60 & 11.56
            & 72.42 & 37.10 & 52.19 & 14.58 
        \\
            & JSL \cite{pratt2020grounded}
            & 39.94 & 31.44 & 18.87 & 24.86 & 9.66
            & 67.60 & 51.88 & 29.39 & 40.60 & 14.72
            & 73.21 & 37.82 & 56.57 & 18.45 
        \\    
        \cline{2-16}
            & GSRTR w/o VE (Ours)
            & 40.61 & 31.87 & 19.01 & 25.21 & 9.69
            & 69.75 & 53.25 & 29.67 & 41.65 & 14.93
            & 72.32 & 36.75 & 56.03 & 18.02        
        \\
            & GSRTR (Ours)
            & \textbf{40.63} & \textbf{32.15} & \textbf{19.28} & \textbf{25.49} & \textbf{10.10}       
            & \textbf{69.81} & \textbf{54.13} & \textbf{31.01} & \textbf{42.50} & \textbf{15.88}
            & \textbf{74.11} & \textbf{39.00} & \textbf{57.45} & \textbf{19.67}      
        \\
        \hline
    \end{tabular}}
\end{table}

\begin{figure}[!t]
    \centering
        \includegraphics[width=1\textwidth]{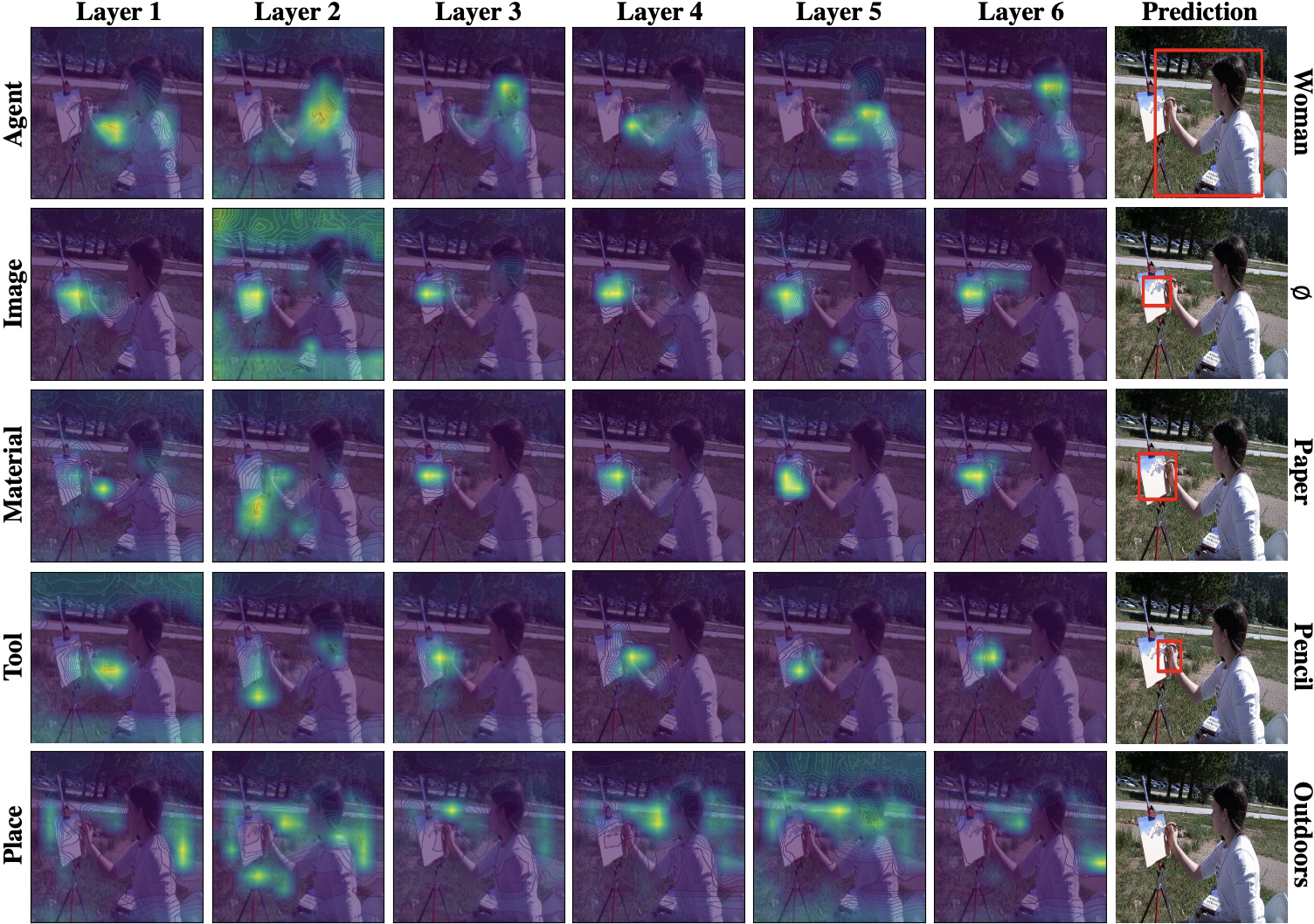}
\caption{
    Role Attention Map on Image Features for a $Sketching$ image from the MHA block in each decoder layer.
    The left labels are the semantic roles of the verb $Sketching$.
    The rightmost column images and labels are predicted bounding boxes and nouns of our model.
}
\label{fig:role_img}
\end{figure}

\noindent
\textbf{Role Attention Map on Image Features:}
In Figure~\ref{fig:role_img}, each column shows the difference of attention maps among semantic roles.
For example, at Layer 6, the role $Agent$ focuses on the woman, and the role $Place$ focuses on the road and yard.
Each row shows the transition of attention maps through the decoder layers.
For example, in the role $Material$, the attention map gradually focuses on the paper in the image through the decoder layers.
It shows that the semantic role queries can focus on the region related to them.

\begin{figure}[!t]
    \centering
    \begin{tabular}{c@{\hskip 0in}c@{\hskip 0in}}
            \includegraphics[height=0.235\textwidth]{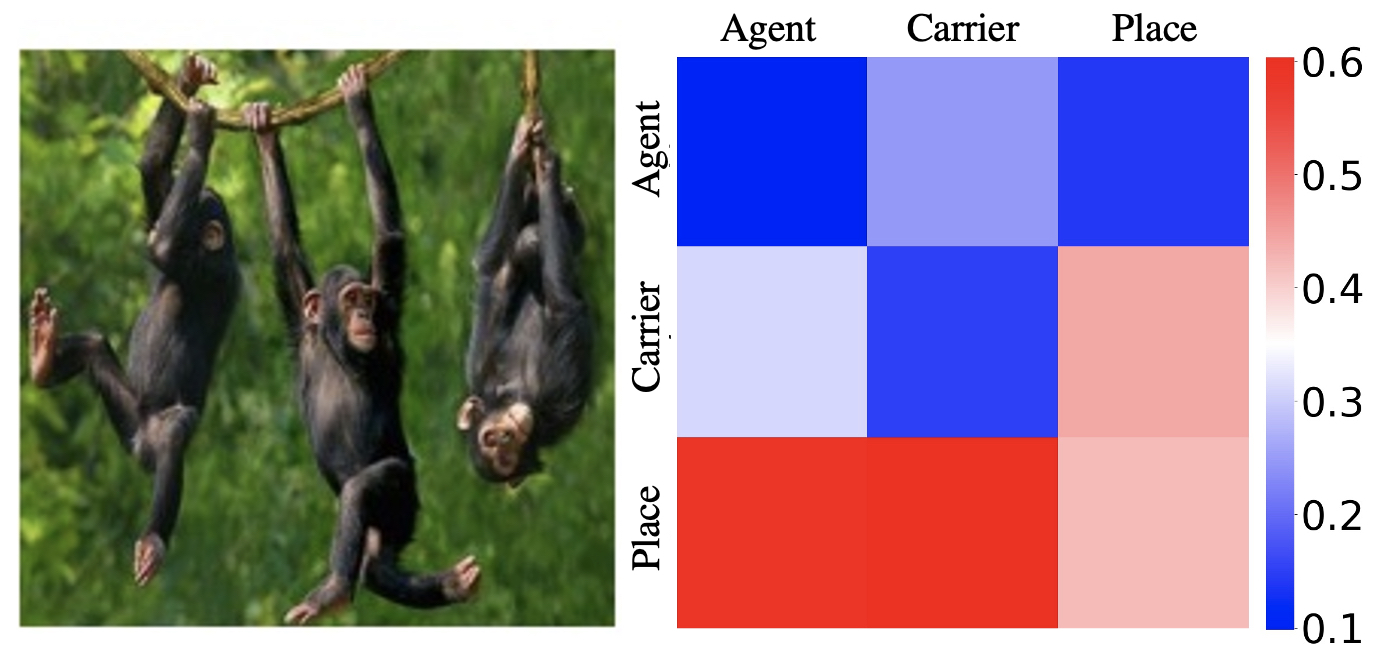}
        &
            \includegraphics[height=0.235\textwidth]{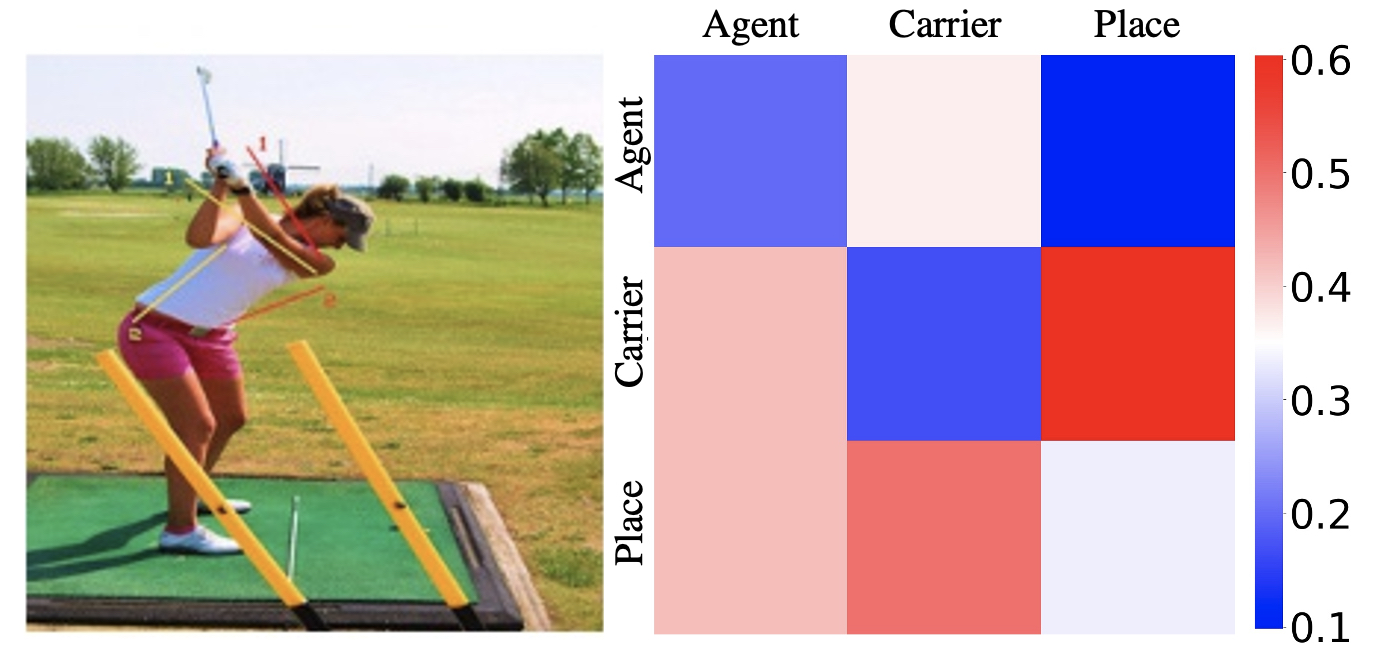}
        \\
       (a)&(b)
    \end{tabular}
\caption{
    Visualization on Role Relations for two \textit{Swinging} images. 
    We visualize the attention scores between semantic role pairs computed in the MHSA block of the last decoder layer. 
    Attention scores are represented as column-wise sum to 1.
}
\label{fig:roles}
\end{figure}

\noindent
\textbf{Visualization on Role Relations:}
In Figure~\ref{fig:roles}, two images show different context for a verb $Swinging$.
The role $Agent$ and $Carrier$ in Fig.~\ref{fig:roles}(a) focus on the role $Place$, \emph{i.e.}, the forest ($Place$) is highly related to the monkey ($Agent$) and the vine ($Carrier$) given the verb $Swinging$.
Meanwhile, the role $Place$ in Fig.~\ref{fig:roles}(b) focuses on the role $Carrier$, \emph{i.e.}, the golf club ($Carrier$) is highly related to the golf course ($Place$) given the verb $Swinging$.
It shows that the relations among roles can be adaptively captured depending on the context of a given image.

\begin{figure}[!t]
    \centering
        \includegraphics[width=\textwidth]{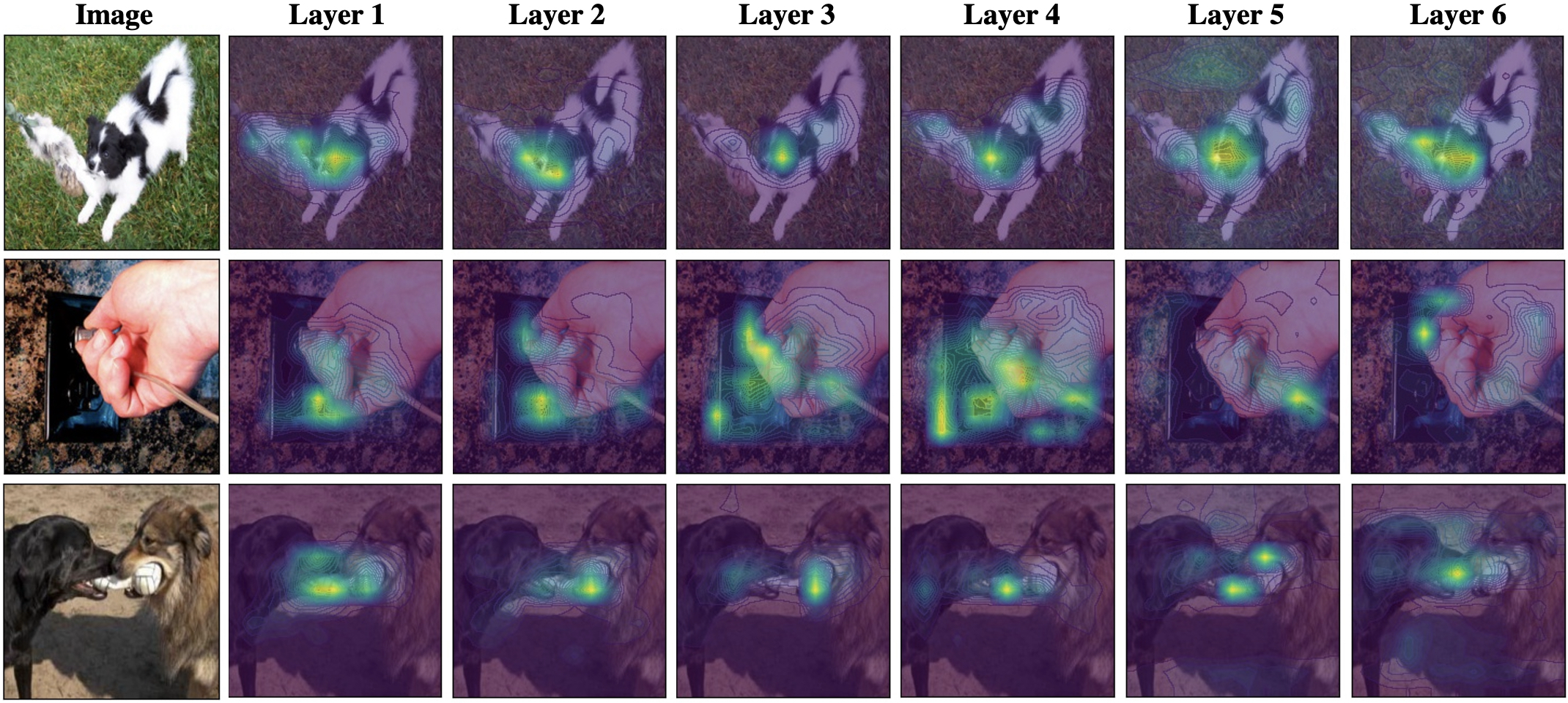}
\caption{
    Verb Token Attention Map on Image Features for three $Tugging$ images.
    Each row consists of an image and attention maps from the MHSA block in each encoder layer.
}
\label{fig:v_img}
\end{figure}

\noindent
\textbf{Verb Token Attention Map on Image Features:}
In Figure~\ref{fig:v_img}, 
the rightmost column shows the semantic regions where the verb token focuses on are similar.
The verb token can capture the key feature (\eg, tugged item) to infer the salient action.
Each row shows the transition of attention maps through the encoder layers,
\eg, focusing on the tugged item gradually.

\section{Discussion}

\begin{figure}[h!]
    \centering
        \includegraphics[width=\textwidth]{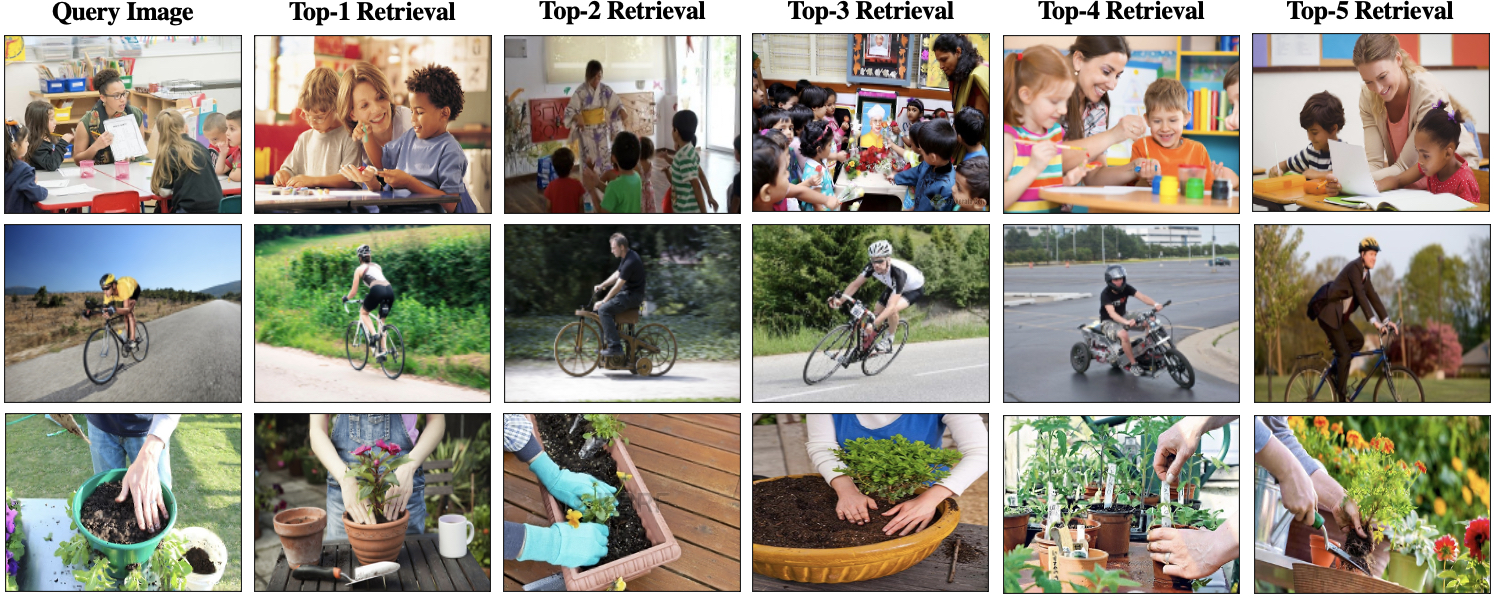}
\caption{
    Grounded-Semantic-Aware Image Retrieval results on the \emph{dev} set. The retrieval results have similar semantics and object arrangements with the query image. In this retrieval, the similarity between two images is computed by the results of verb prediction and grounded noun prediction as in~\cite{pratt2020grounded}. 
}
\label{fig:retrieval}
\end{figure}

\noindent There have been many studies in image retrieval by computing the similarities between the visual representations of images.
But, they do not work well for getting the retrieval results which have similar situations with respect to semantics or object arrangements.
Grounded-Semantic-Aware Image Retrieval enables image retrieval in the aspects of main activity and key objects with their arrangements, as shown in Figure~\ref{fig:retrieval}.
This retrieval uses the results of verb prediction and grounded noun prediction instead of visual representations.
The predictions of main activity (\textit{verb}) and entities (\textit{nouns}) enable image retrieval for similar semantics, and the predictions of entity locations enable image retrieval for similar object arrangements.
In this retrieval, we compute the  $\mathrm{GrSitSim}(I, J)$~\cite{pratt2020grounded} as similarity score function between image $I$ and $J$.
For an image $I$, we compute the top-5 verb predictions $\hat{v}_1^{I}$, ..., $\hat{v}_5^{I}$. For each verb prediction $\hat{v}_i^{I}$, we predict nouns $\hat{n}_{i,1}^{I}$, ..., $\hat{n}_{i,|\mathcal{R}_{\hat{v}_{i}^{I}}|}^{I}$ and bounding boxes $\mathbf{\hat{b}}_{i,1}^{I}$, ..., $\mathbf{\hat{b}}_{i,|\mathcal{R}_{\hat{v}_{i}^{I}}|}^{I}$.
Note that we ignore the predicted bounding box if its existence probability is less than 0.5. We calculate the similarity between two images $I$ and $J$ as follows:

\begin{align}
    \label{eq:grsitsim}
    \nonumber \mathrm{Gr}&\mathrm{SitSim}(I, J)
    \\
    &= \max\left\{ \frac{ \mathbbm 1_{[\hat{v}_i^{I} = \hat{v}_j^{J}]}}{2 \cdot i\cdot j \cdot \vert \mathcal R_{\hat{v}_i^I} \vert} \mathlarger{\sum_{k=1}^{\vert \mathcal{R}_{\hat{v}_{i}^{I}} \vert}}
    \mathbbm 1_{[\hat{n}_{i,k}^{I} = \hat{n}_{j,k}^{J}]}\cdot
    \left(1+\mathrm{IoU}(\mathbf{\hat{b}}_{i,k}^{I}, \mathbf{\hat{b}}_{j,k}^{J})\right)  
    \Bigg\vert  1 \leq i,j \leq 5\right\}.
\end{align}

\noindent $\mathrm{GrSitSim}(I, J)$ is computed by the results of verb prediction and grounded noun prediction for image $I$ and $J$. 
The similarity is not zero when at least one verb is shared in the top-5 verb predictions for image $I$ and $J$. 
The similarity is maximized when the top-1 verb predictions and noun predictions of two images are same, and the sizes and locations of predicted bounding boxes are same. For this reason, we can get the retrieval result which has similar semantics and object arrangements in Grounded-Semantic-Aware Image Retrieval. 
Thus, we can apply this image retrieval to the applications where semantics and object arrangements are important, \eg, search engine using semantics and object arrangements of images.

Grounded Situation Recognition models produce complete predictions with respect to the semantic roles corresponding to a verb.
Thus, the models can answer the following questions more strictly, ``What is the main activity'' (\textit{verb}), ``Who is participating in the main activity'' (role \textit{Agent}), ``What does the actor use in the main activity'' (role \textit{Tool}), 
``Where is the actor in the image'' (entity location of role \textit{Agent}), etc.
For this reason, the models are useful for predetermined questions on situations.
Taking advantages of these properties, we can apply the models for industry such as unmanned surveillance system or service robot.

\section{Conclusion}
We propose the first Transformer architecture for GSR, which achieves the state-of-the-art accuracy on every evaluation metric.
Our model, GSRTR, can capture high-level semantic feature, and flexibly deal with the complicated and image-dependent role relations. We perform extensive experiments and qualitatively illustrate the effectiveness of our method.

\vspace{3mm}
{\noindent \textbf{Acknowledgement:} 
This work was supported by the NRF grant and the IITP grant funded by Ministry of Science and ICT, Korea
(No.2019-0-01906 Artificial Intelligence Graduate School Program--POSTECH,
 NRF-2021R1A2C3012728--50\%, 
 IITP-2020-0-00842--50\%).}
\bibliography{bmvc_arxiv_latex_cleaner}

\begin{thebibliography}{34}
\providecommand{\natexlab}[1]{#1}
\providecommand{\url}[1]{\texttt{#1}}
\expandafter\ifx\csname urlstyle\endcsname\relax
  \providecommand{\doi}[1]{doi: #1}\else
  \providecommand{\doi}{doi: \begingroup \urlstyle{rm}\Url}\fi

\bibitem[Carion et~al.(2020)Carion, Massa, Synnaeve, Usunier, Kirillov, and
  Zagoruyko]{carion2020end}
Nicolas Carion, Francisco Massa, Gabriel Synnaeve, Nicolas Usunier, Alexander
  Kirillov, and Sergey Zagoruyko.
\newblock {End-to-End Object Detection with Transformers}.
\newblock In \emph{Proceedings of the European Conference on Computer Vision
  (ECCV)}, pages 213--229, 2020.

\bibitem[Cooray et~al.(2020)Cooray, Cheung, and Lu]{cooray2020attention}
Thilini Cooray, Ngai-Man Cheung, and Wei Lu.
\newblock {Attention-Based Context Aware Reasoning for Situation Recognition}.
\newblock In \emph{Proceedings of the IEEE/CVF Conference on Computer Vision
  and Pattern Recognition (CVPR)}, pages 4736--4745, 2020.

\bibitem[Dosovitskiy et~al.(2021)Dosovitskiy, Beyer, Kolesnikov, Weissenborn,
  Zhai, Unterthiner, Dehghani, Minderer, Heigold, Gelly, Uszkoreit, and
  Houlsby]{dosovitskiy2021an}
Alexey Dosovitskiy, Lucas Beyer, Alexander Kolesnikov, Dirk Weissenborn,
  Xiaohua Zhai, Thomas Unterthiner, Mostafa Dehghani, Matthias Minderer, Georg
  Heigold, Sylvain Gelly, Jakob Uszkoreit, and Neil Houlsby.
\newblock {An Image is Worth 16x16 Words: Transformers for Image Recognition at
  Scale}.
\newblock In \emph{International Conference on Learning Representations
  (ICLR)}, 2021.

\bibitem[Fillmore et~al.(2003)Fillmore, Johnson, and
  Petruck]{fillmore2003background}
Charles~J. Fillmore, Christopher~R. Johnson, and Miriam~R.L. Petruck.
\newblock {Background to Framenet}.
\newblock \emph{International Journal of Lexicography}, 16\penalty0
  (3):\penalty0 235--250, 2003.

\bibitem[Glorot and Bengio(2010)]{xavier2010init}
Xavier Glorot and Yoshua Bengio.
\newblock {Understanding the difficulty of training deep feedforward neural
  networks}.
\newblock In \emph{Proceedings of the Thirteenth International Conference on
  Artificial Intelligence and Statistics}, pages 249--256, 2010.

\bibitem[Gong et~al.(2014)Gong, Wang, Guo, and Lazebnik]{gong2014multi}
Yunchao Gong, Liwei Wang, Ruiqi Guo, and Svetlana Lazebnik.
\newblock {Multi-scale Orderless Pooling of Deep Convolutional Activation
  Features}.
\newblock In \emph{Proceedings of the European Conference on Computer Vision
  (ECCV)}, pages 392--407, 2014.

\bibitem[{He, Kaiming and Zhang, Xiangyu and Ren, Shaoqing and Sun,
  Jian}(2016)]{resnet}
{He, Kaiming and Zhang, Xiangyu and Ren, Shaoqing and Sun, Jian}.
\newblock {Deep Residual Learning for Image Recognition}.
\newblock In \emph{Proceedings of the IEEE Conference on Computer Vision and
  Pattern Recognition (CVPR)}, pages 770--778, 2016.

\bibitem[Huang et~al.(2019)Huang, Wang, Chen, and Wei]{huang2019attention}
Lun Huang, Wenmin Wang, Jie Chen, and Xiao-Yong Wei.
\newblock {Attention on Attention for Image Captioning}.
\newblock In \emph{Proceedings of the IEEE/CVF International Conference on
  Computer Vision (ICCV)}, pages 4634--4643, 2019.

\bibitem[Li et~al.(2017)Li, Tapaswi, Liao, Jia, Urtasun, and
  Fidler]{li2017situation}
Ruiyu Li, Makarand Tapaswi, Renjie Liao, Jiaya Jia, Raquel Urtasun, and Sanja
  Fidler.
\newblock {Situation Recognition with Graph Neural Network}.
\newblock In \emph{Proceedings of the IEEE International Conference on Computer
  Vision (ICCV)}, pages 4173--4182, 2017.

\bibitem[Li et~al.(2016)Li, Tarlow, Brockschmidt, and Zemel]{li2016gated}
Yujia Li, Daniel Tarlow, Marc Brockschmidt, and Richard Zemel.
\newblock {Gated Graph Sequence Neural Networks}.
\newblock In \emph{International Conference on Learning Representations
  (ICLR)}, 2016.

\bibitem[Lin et~al.(2017)Lin, Dollar, Girshick, He, Hariharan, and
  Belongie]{lin2017_fpn}
Tsung-Yi Lin, Piotr Dollar, Ross Girshick, Kaiming He, Bharath Hariharan, and
  Serge Belongie.
\newblock {Feature Pyramid Networks for Object Detection}.
\newblock In \emph{Proceedings of the IEEE Conference on Computer Vision and
  Pattern Recognition (CVPR)}, pages 2117--2125, 2017.

\bibitem[Liu et~al.(2021)Liu, Yan, Mortazavi, and Bhanu]{liu2021fully}
Hengyue Liu, Ning Yan, Masood Mortazavi, and Bir Bhanu.
\newblock {Fully Convolutional Scene Graph Generation}.
\newblock In \emph{Proceedings of the IEEE/CVF Conference on Computer Vision
  and Pattern Recognition (CVPR)}, pages 11546--11556, 2021.

\bibitem[Liu et~al.(2019)Liu, Tian, and Xu]{liu2019novel}
Shaopeng Liu, Guohui Tian, and Yuan Xu.
\newblock {A novel scene classification model combining ResNet based transfer
  learning and data augmentation with a filter}.
\newblock \emph{Neurocomputing}, 338:\penalty0 191--206, 2019.

\bibitem[Loshchilov and Hutter(2019)]{loshchilov2018decoupled}
Ilya Loshchilov and Frank Hutter.
\newblock {Decoupled Weight Decay Regularization}.
\newblock In \emph{International Conference on Learning Representations
  (ICLR)}, 2019.

\bibitem[Mallya and Lazebnik(2017)]{mallya2017recurrent}
Arun Mallya and Svetlana Lazebnik.
\newblock {Recurrent Models for Situation Recognition}.
\newblock In \emph{Proceedings of the IEEE International Conference on Computer
  Vision (ICCV)}, pages 455--463, 2017.

\bibitem[Pham et~al.(2021)Pham, Dai, Xie, and Le]{pham2021meta}
Hieu Pham, Zihang Dai, Qizhe Xie, and Quoc~V. Le.
\newblock {Meta Pseudo Labels}.
\newblock In \emph{Proceedings of the IEEE/CVF Conference on Computer Vision
  and Pattern Recognition (CVPR)}, pages 11557--11568, 2021.

\bibitem[Pratt et~al.(2020)Pratt, Yatskar, Weihs, Farhadi, and
  Kembhavi]{pratt2020grounded}
Sarah Pratt, Mark Yatskar, Luca Weihs, Ali Farhadi, and Aniruddha Kembhavi.
\newblock {Grounded Situation Recognition}.
\newblock In \emph{Proceedings of the European Conference on Computer Vision
  (ECCV)}, pages 314--332, 2020.

\bibitem[{Rezatofighi, Hamid and Tsoi, Nathan and Gwak, JunYoung and Sadeghian,
  Amir and Reid, Ian and Savarese, Silvio}(2019)]{rezatofighi2019generalized}
{Rezatofighi, Hamid and Tsoi, Nathan and Gwak, JunYoung and Sadeghian, Amir and
  Reid, Ian and Savarese, Silvio}.
\newblock {Generalized Intersection Over Union: A Metric and a Loss for
  Bounding Box Regression}.
\newblock In \emph{Proceedings of the IEEE/CVF Conference on Computer Vision
  and Pattern Recognition (CVPR)}, pages 658--666, 2019.

\bibitem[Safaei and Foroosh(2019)]{safaei2019still}
Marjaneh Safaei and Hassan Foroosh.
\newblock {Still Image Action Recognition by Predicting Spatial-Temporal Pixel
  Evolution}.
\newblock In \emph{2019 IEEE Winter Conference on Applications of Computer
  Vision (WACV)}, pages 111--120, 2019.
\newblock \doi{10.1109/WACV.2019.00019}.

\bibitem[Suhail and Sigal(2019)]{suhail2019mixture}
Mohammed Suhail and Leonid Sigal.
\newblock {Mixture-Kernel Graph Attention Network for Situation Recognition}.
\newblock In \emph{Proceedings of the IEEE/CVF International Conference on
  Computer Vision (ICCV)}, pages 10363--10372, 2019.

\bibitem[Szegedy et~al.(2016)Szegedy, Vanhoucke, Ioffe, Shlens, and
  Wojna]{szegedy2016rethinking}
Christian Szegedy, Vincent Vanhoucke, Sergey Ioffe, Jon Shlens, and Zbigniew
  Wojna.
\newblock {Rethinking the Inception Architecture for Computer Vision}.
\newblock In \emph{Proceedings of the IEEE Conference on Computer Vision and
  Pattern Recognition (CVPR)}, pages 2818--2826, 2016.

\bibitem[Vaswani et~al.(2017)Vaswani, Shazeer, Parmar, Uszkoreit, Jones, Gomez,
  Kaiser, and Polosukhin]{vaswani2017attention}
Ashish Vaswani, Noam Shazeer, Niki Parmar, Jakob Uszkoreit, Llion Jones,
  Aidan~N Gomez, \L{}ukasz Kaiser, and Illia Polosukhin.
\newblock {Attention is All you Need}.
\newblock In \emph{Advances in Neural Information Processing Systems (NIPS)},
  2017.

\bibitem[Vinyals et~al.(2015)Vinyals, Toshev, Bengio, and
  Erhan]{vinyals2015show}
Oriol Vinyals, Alexander Toshev, Samy Bengio, and Dumitru Erhan.
\newblock {Show and Tell: A Neural Image Caption Generator}.
\newblock In \emph{Proceedings of the IEEE Conference on Computer Vision and
  Pattern Recognition (CVPR)}, pages 3156--3164, 2015.

\bibitem[Wang et~al.(2021)Wang, Xu, Wang, Shen, Cheng, Shen, and
  Xia]{wang2021end}
Yuqing Wang, Zhaoliang Xu, Xinlong Wang, Chunhua Shen, Baoshan Cheng, Hao Shen,
  and Huaxia Xia.
\newblock {End-to-End Video Instance Segmentation With Transformers}.
\newblock In \emph{Proceedings of the IEEE/CVF Conference on Computer Vision
  and Pattern Recognition (CVPR)}, pages 8741--8750, 2021.

\bibitem[Xiong et~al.(2020)Xiong, Yang, He, Zheng, Zheng, Xing, Zhang, Lan,
  Wang, and Liu]{xiong2020layer}
Ruibin Xiong, Yunchang Yang, Di~He, Kai Zheng, Shuxin Zheng, Chen Xing,
  Huishuai Zhang, Yanyan Lan, Liwei Wang, and Tieyan Liu.
\newblock {On Layer Normalization in the Transformer Architecture}.
\newblock In \emph{International Conference on Machine Learning (ICML)}, pages
  10524--10533. PMLR, 2020.

\bibitem[Xu et~al.(2017)Xu, Zhu, Choy, and Fei-Fei]{xu2017scene}
Danfei Xu, Yuke Zhu, Christopher~B Choy, and Li~Fei-Fei.
\newblock {Scene Graph Generation by Iterative Message Passing}.
\newblock In \emph{Proceedings of the IEEE Conference on Computer Vision and
  Pattern Recognition (CVPR)}, pages 5410--5419, 2017.

\bibitem[Yang et~al.(2018)Yang, Lu, Lee, Batra, and Parikh]{yang2018graph}
Jianwei Yang, Jiasen Lu, Stefan Lee, Dhruv Batra, and Devi Parikh.
\newblock {Graph R-CNN for Scene Graph Generation}.
\newblock In \emph{Proceedings of the European Conference on Computer Vision
  (ECCV)}, pages 670--685, 2018.

\bibitem[Yatskar et~al.(2016)Yatskar, Zettlemoyer, and
  Farhadi]{yatskar2016situation}
Mark Yatskar, Luke Zettlemoyer, and Ali Farhadi.
\newblock {Situation Recognition: Visual Semantic Role Labeling for Image
  Understanding}.
\newblock In \emph{Proceedings of the IEEE Conference on Computer Vision and
  Pattern Recognition (CVPR)}, pages 5534--5542, 2016.

\bibitem[Yatskar et~al.(2017)Yatskar, Ordonez, Zettlemoyer, and
  Farhadi]{yatskar2017commonly}
Mark Yatskar, Vicente Ordonez, Luke Zettlemoyer, and Ali Farhadi.
\newblock {Commonly Uncommon: Semantic Sparsity in Situation Recognition}.
\newblock In \emph{Proceedings of the IEEE Conference on Computer Vision and
  Pattern Recognition (CVPR)}, pages 7196--7205, 2017.

\bibitem[You et~al.(2016)You, Jin, Wang, Fang, and Luo]{you2016image}
Quanzeng You, Hailin Jin, Zhaowen Wang, Chen Fang, and Jiebo Luo.
\newblock {Image Captioning with Semantic Attention}.
\newblock In \emph{Proceedings of the IEEE Conference on Computer Vision and
  Pattern Recognition (CVPR)}, pages 4651--4659, 2016.

\bibitem[Zhang et~al.(2021)Zhang, Gupta, and Zisserman]{zhang2021temporal}
Chuhan Zhang, Ankush Gupta, and Andrew Zisserman.
\newblock {Temporal Query Networks for Fine-grained Video Understanding}.
\newblock In \emph{Proceedings of the IEEE/CVF Conference on Computer Vision
  and Pattern Recognition (CVPR)}, pages 4486--4496, 2021.

\bibitem[Zhao et~al.(2017)Zhao, Ma, and You]{zhao2017single}
Zhichen Zhao, Huimin Ma, and Shaodi You.
\newblock {Single Image Action Recognition using Semantic Body Part Actions}.
\newblock In \emph{Proceedings of the IEEE International Conference on Computer
  Vision (ICCV)}, pages 3391--3399, 2017.

\bibitem[Zhou et~al.(2014)Zhou, Lapedriza, Xiao, Torralba, and
  Oliva]{zhou2014learning}
Bolei Zhou, Agata Lapedriza, Jianxiong Xiao, Antonio Torralba, and Aude Oliva.
\newblock {Learning Deep Features for Scene Recognition using Places Database}.
\newblock In \emph{Advances in Neural Information Processing Systems (NIPS)},
  2014.

\bibitem[Zou et~al.(2021)Zou, Wang, Hu, Liu, Wu, Zhao, Li, Zhang, Zhang, Wei,
  et~al.]{zou2021end}
Cheng Zou, Bohan Wang, Yue Hu, Junqi Liu, Qian Wu, Yu~Zhao, Boxun Li, Chenguang
  Zhang, Chi Zhang, Yichen Wei, et~al.
\newblock {End-to-End Human Object Interaction Detection with HOI Transformer}.
\newblock In \emph{Proceedings of the IEEE/CVF Conference on Computer Vision
  and Pattern Recognition (CVPR)}, pages 11825--11834, 2021.

\end{thebibliography}
\clearpage
\setcounter{section}{0}
\setcounter{figure}{0}
\setcounter{table}{0}
\renewcommand\thesection{\Alph{section}}
\renewcommand\thesubsection{\thesection.\arabic{subsection}}
\renewcommand\thefigure{A\arabic{figure}}

\section{Appendix}
This provides more details of our model, further analyses on it, additional ablation studies and experimental results.
Section~\ref{supp:arch} describes the transformer architecture of GSRTR in detail, Section~\ref{supp:abla} performs the ablation studies on GSRTR, and Section~\ref{supp:pred} provides more qualitative examples of the total prediction of GSRTR. Finally, a more thorough qualitative analysis on attention of GSRTR is illustrated in Section~\ref{supp:viz}.

\subsection{Detailed Transformer Architecture}
\label{supp:arch}

\begin{figure}[!htb]
\centering
    \includegraphics[height=0.8\textheight]{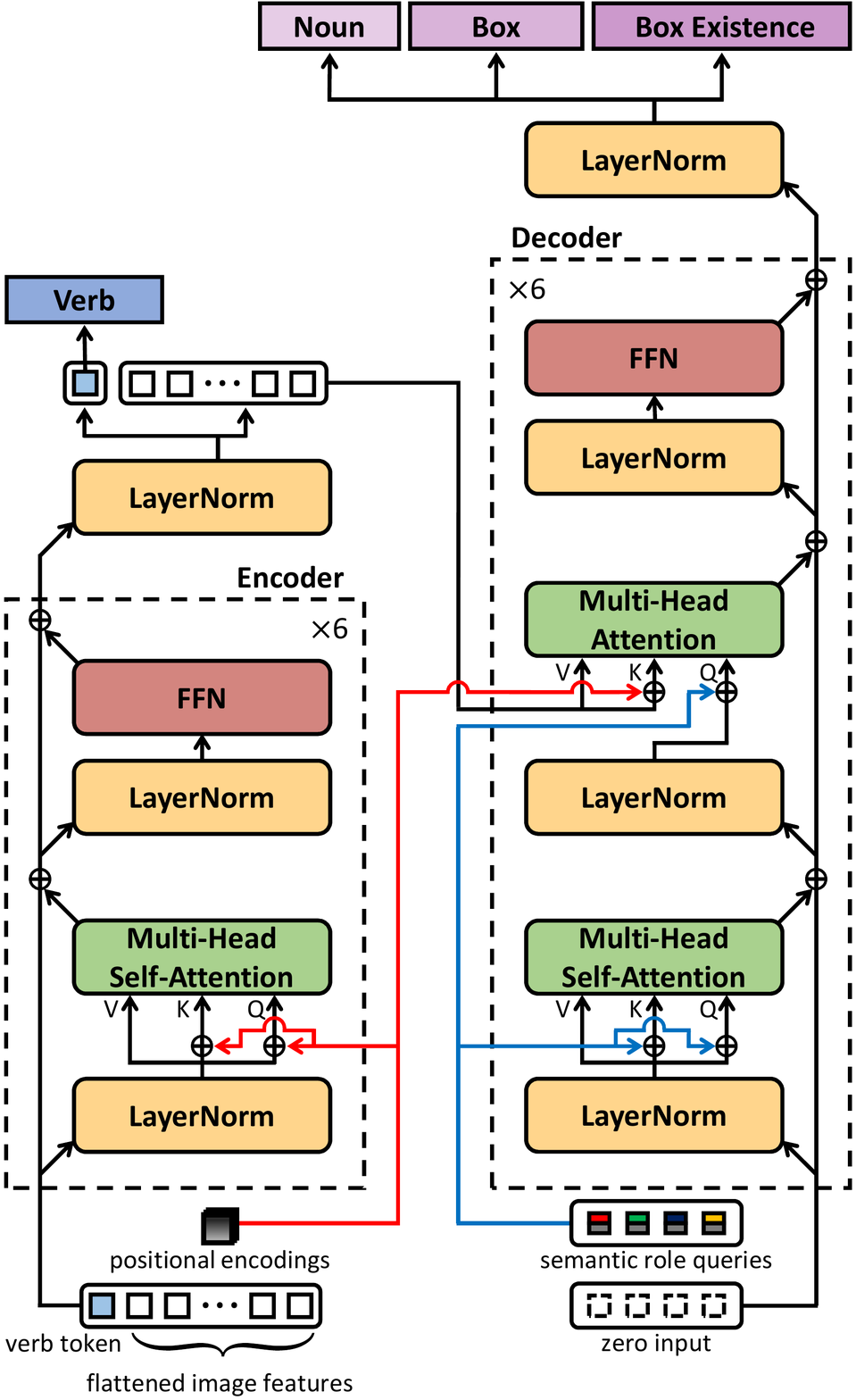}
\caption{
The detailed transformer architecture of GSRTR. 
A verb token and flattened image features are used for the first encoder layer input (black line in Encoder).
Zero input is used for the first decoder layer input (black line in Decoder).
Positional encodings are added to the keys and queries of the MHSA block in each encoder layer and the keys of the MHA block in each decoder layer (red line).
Semantic role queries are added to the keys and queries of the MHSA block in each decoder layer and the queries of the MHA block in each decoder layer (blue line).
We omit Dropout in this diagram.}
\label{fig:supp_detailed}
\end{figure}

\noindent
\textbf{Transformer Encoder-Decoder:}
The detailed transformer architecture of GSRTR is given in Figure~\ref{fig:supp_detailed}.
The encoder takes as input a verb token and flattened image features, and then produces a verb feature and image features.
Along with image features given by the encoder, the decoder takes as input semantic role queries, and then produces output features corresponding to the semantic roles.
The encoder is a stack of six encoder layers and the decoder is a stack of six decoder layers.
Each encoder layer consists of a Multi-Head Self-Attention (MHSA) block and a Feed-Forward Network (FFN) block.
Each decoder layer consists of a MHSA block, a Multi-Head Attention (MHA) block, and a FFN block.
We use Pre-Layer Normalization (Pre-LN)~\cite{xiong2020layer}, \emph{i.e.}, LayerNorm is used before each MHSA block, MHA block, and FFN block, and also before the verb feature and before the decoder output features corresponding to the semantic roles.
The skip connection, using $0.15$ dropout rate, is given by:
\begin{align}\tag{A.1}
    \mathbf x+\mathrm{Dropout}\left(\mathrm{Block}(\mathrm{LayerNorm}\left(\mathbf x)\right)\right),
\end{align}
where $\mathbf{x}\in\mathbb R^{d}$ and Block denotes one of the MHSA block, MHA block, and FFN block. 
Note that we use $d=512$.
The FFN block is 2 fully-connected layers with ReLU activation function and 2048 hidden dimensions, using $0.15$ dropout rate, and it is given by:
\begin{align}\tag{A.2}
    \mathrm{FFN}(\mathbf x) = W_2\left(\mathrm{Dropout}\left(\max(W_1 \mathbf x + \mathbf{b}_1, \mathbf{0})\right)\right) + \mathbf{b}_2,
\end{align}
where $\mathbf{x}\in\mathbb R^d$, $W_1\in\mathbb R^{2048\times d}$, $\mathbf{b}_1\in \mathbb R^{2048}$, $W_2\in\mathbb R^{d\times 2048}$, and $\mathbf{b}_2 \in \mathbb R^{d}$.
We use Xavier initialization~\cite{xavier2010init} for the learnable parameters in the encoder and decoder.

\vspace{2mm}
\noindent
\textbf{Multi-Head Attention:}
MHA takes as input a query sequence $X_Q\in\mathbb R^{d \times n_Q}$ and a key-value sequence $X_{KV} \in \mathbb R^{d \times n_{KV}}$, where $n_Q$ denotes the query sequence length and $n_{KV}$ denotes the key-value sequence length.
MHSA corresponds to the case when the query sequence is same with the key-value sequence in MHA, \emph{i.e.,} when $X_Q = X_{KV}$ in MHA.
MHA is formulated as:
\begin{align}\tag{A.3}
    \mathrm{MHA} (X_Q, X_{KV}) &= W_O \left[\mathrm{Head}^1\left(X_Q, X_{KV}\right); \cdots; \mathrm{Head}^H\left(X_Q, X_{KV}\right)\right],
\end{align}
where $H$ is the number of heads, $[;]$ denotes a concatenation and $W_O\in \mathbb R^{d\times d}$ denotes an output projection.
Note that we use $H=8$.
$\mathrm{Head}^m$ denotes each attention function with linear projections for $m=1,\cdots,H$, and it is given by:
\begin{align}\tag{A.4}
    \mathrm{Head}^m(X_Q, X_{KV}) &= \mathrm{Attn}\left(W_Q^m X_Q, W_K^m X_{KV}, W_V^m X_{KV}\right),
\end{align}
where $W_Q^m, W_K^m, W_V^m\in \mathbb R^{d^\prime \times d}$ denotes linear projection of $m^\mathrm{th}$ head for key, query, and value, respectively.
The linear projection matrices are learnable parameters, which are not shared across the MHA and MHSA blocks in the encoder and decoder layers.
Note that we use $d^\prime = 64$, where $d^\prime = \frac{d}{H}$.
$\mathrm{Attn}$ denotes an attention function which transforms a query sequence $Q\in\mathbb R^{d^\prime \times n_Q}$ into an output sequence, whose element is a weighted sum of a value sequence $V\in\mathbb R^{d^\prime \times n_{KV}}$.
For $i^\mathrm{th}$ query $\mathbf{q}_i\in\mathbb R^{d^\prime}$, 
each weight of the sum 
is computed by a softmax function (\emph{i.e.}, $\mathrm{Softmax}$) after a scaled dot-product between the $i^\mathrm{th}$ query $\mathbf{q}_i$ and a key sequence $K\in\mathbb R^{d^\prime \times n_{KV}}$.
In other words, the $i^\mathrm{th}$ element of the attention function output from the query sequence $Q$, key sequence $K$, and value sequence $V$ is given by:

\begin{align}\tag{A.5}
    \mathrm{Attn}_i (Q, K, V) = \sum_{j} \mathrm{Softmax}_j\left(\frac{1}{\sqrt{d^\prime}} \mathbf{q}_i K\right) \mathbf{v}_j,
\end{align}
where $\mathrm{Softmax}_j$ denotes the $j^\mathrm{th}$ output of the softmax function and $\mathbf{v}_j\in\mathbb R^{d^\prime}$ denotes the $j^\mathrm{th}$ value.

\vspace{2mm}
\noindent
\textbf{The MHSA block in the encoder:}
The encoder takes as input a verb token and flattened image features.
The positional encodings $P\in\mathbb R^{d \times hw}$ are used,
where $hw$ denotes the length of flattened image features.
The positional encodings $P$ are 2D learnable embeddings, and they are used at the attention function of each MHSA block in the encoder.
To be specific, the positional encodings are added to the corresponding image features, which are used as the key and query inputs at the attention function.
For the verb token, we append zero to the positional encodings, leading to  $P^\prime \in \mathbb R^{d \times (1+hw)}$.
As a result, the positional encodings $P^\prime$ are added to the key and query inputs of the attention function in each MHSA block of the encoder. Thus, the $m^\mathrm{th}$ attention function in each MHSA block of the encoder is given by:
\begin{align}\tag{A.6}
    \mathrm{Head}^m(X_Q, X_{KV}) &= \mathrm{Attn}\left(W_Q^m \left(X_Q + P^\prime\right), W_K^m \left(X_{KV}+P^\prime\right), W_V^m X_{KV}\right),
\end{align}
where $X_Q=X_{KV}$ and $X_Q\in \mathbb R^{d \times (1+hw)}$.

\vspace{2mm}
\noindent
\textbf{The MHSA and MHA blocks in the decoder:}
Along with the image features given by the encoder, the decoder takes as input a sequence of the semantic role queries.
Additionally to Section~\ref{sec:noun}, each semantic role query $\mathbf{w}_{(v, r)}$ per semantic role $r\in\mathcal R_v$ can formulate a sequence with arbitrary role orders, leading to the semantic role query sequence $S_v\in\mathbb R^{d\times \vert \mathcal R_v \vert}$.
Note that the initial decoder input is set to zero.
In each MHSA block of the decoder, the semantic role query sequence $S_v$ is added to the query and key inputs of the attention function.
In other words, the $m^\mathrm{th}$ attention function in each MHSA block of the decoder is given by:
\begin{align}\tag{A.7}
    \mathrm{Head}^m\left(X_Q, X_{KV}\right) &= \mathrm{Attn}\left(W_Q^m \left(X_Q + S_v\right), W_K^m (X_{KV}+S_v), W_V^m X_{KV}\right),
\end{align}
where $X_Q=X_{KV}$ and $X_Q\in \mathbb R^{d \times \vert \mathcal R_v \vert}$. 
In each MHA block of the decoder, the semantic role query sequence $S_v$ are added to the query inputs of the attention function, and positional encodings $P$ are added to the key inputs of the attention function.
In other words, the $m^\mathrm{th}$ attention function in each MHA block of the decoder is given by:
\begin{align}\tag{A.8}
    \mathrm{Head}^m\left(X_Q, X_{KV}\right) &= \mathrm{Attn}\left(W_Q^m \left(X_Q + S_v\right), W_K^m \left(X_{KV}+P\right), W_V^m X_{KV}\right),
\end{align}
where $X_Q\in \mathbb R^{d \times \vert \mathcal R_v \vert}$ 
and
$X_{KV}\in \mathbb R^{d \times hw}$.

\clearpage
\subsection{Ablation Studies}
\label{supp:abla}
\renewcommand\thetable{A\arabic{table}}

\begin{table}[!t]
    \centering
    \caption{
        Ablation studies on our model (GSRTR).
    }
    \resizebox{\textwidth}{!}{
        \begin{tabular}{l|l|ccccc|ccccc|cccc}
        \hline
        \multicolumn{2}{c|}{}
            & \multicolumn{5}{c|}{top-1 predicted verb}
            & \multicolumn{5}{c|}{top-5 predicted verbs}
            & \multicolumn{4}{c}{ground-truth verb}  
        \\
        \hline
            &  
            &       &       &       & grnd & grnd
            &       &       &       & grnd & grnd
            &       &       & grnd & grnd  
        \\
        set & model 
            & verb & value & value-all & value & value-all
            & verb & value & value-all & value & value-all
            & value & value-all & value & value-all
        \\
        \hline
        \hline
        \multirow{4}{*}{dev} & GSRTR w/ 4 layers
            & 40.26 & 31.88 & 19.20 & 25.44 & 10.20
            & 69.34 & 53.52 & 30.33 & 42.29 & 15.69
            & 74.09 & 38.88 & 57.97 & 19.75     
        \\
            & GSRTR w/ 8 layers
            & 40.49 & 32.10 & 19.46 & 25.69 & 10.39      
            & 69.11 & 53.34 & 30.62 & 42.35 & 15.88     
            & 74.07 & 39.12 & 58.27 & 19.92     
        \\
            & GSRTR w/ Post-LN
            & 40.18      & 31.50      & 18.54      & 25.20       & 9.89 
            & 68.82      & 52.72      & 29.30      & 41.79       & 15.27     
            & 73.30      & 37.60      & 57.50      & 19.34      
        \\
        \cline{2-16}
            & GSRTR
            & \textbf{41.06} & \textbf{32.52} & \textbf{19.63} & \textbf{26.04} & \textbf{10.44}
            & \textbf{69.46} & \textbf{53.69} & \textbf{30.66} & \textbf{42.61} & \textbf{15.98}
            & \textbf{74.27} & \textbf{39.24} & \textbf{58.33} & \textbf{20.19} 
        \\
        \hline
        \hline
        \multirow{4}{*}{test} & GSRTR w/ 4 layers
            & \textbf{40.87}      & \textbf{32.21}       & 19.13       & 25.35       & 9.83      
            & \textbf{69.87}      & 53.78      & 30.25       & 41.97       & 15.22      
            & 73.89      & 38.42      & 57.00       & 18.88      
        \\
            & GSRTR w/ 8 layers
            & 40.83      & 32.20      & 19.17       & \textbf{25.49}      & 10.03     
            & 69.47      & 53.40      & 30.07       & 41.99      & 15.35      
            & 73.75      & 38.54      & 57.20       & 19.19      
        \\
            & GSRTR w/ Post-LN
            & 40.31      & 31.72      & 18.69       & 25.03      &  9.56    
            & 69.86      & 53.57      & 29.89       & 41.99      &  15.14    
            & 73.33      & 37.76      & 56.70       & 18.78     
        \\
        \cline{2-16}
            & GSRTR
            & 40.63 & 32.15 & \textbf{19.28} & \textbf{25.49} & \textbf{10.10}       
            & 69.81 & \textbf{54.13} & \textbf{31.01} & \textbf{42.50} & \textbf{15.88}
            & \textbf{74.11} & \textbf{39.00} & \textbf{57.45} & \textbf{19.67}      
        \\
        \hline
    \end{tabular}}
    \label{table:abla}
\end{table}
We study the effect on the number of layers and the location of LayerNorm in GSRTR. 
Our experiments are evaluated on the \emph{dev} and \emph{test} splits of SWiG dataset~\cite{pratt2020grounded}, and the results are compared with 
the proposed model and setting in Section~\ref{exp:detail}.

The effect on the number of layers in the encoder and decoder is shown at the first and second row of each set in Table~\ref{table:abla}.
GSRTR w/ 4 layers denotes that each of the transformer encoder and decoder has four layers, and GSRTR w/ 8 layers denotes that each has eight layers.
In ground-truth verb setting, the noun and grounded noun accuracies of both models decrease.
The top-1 predicted verb and top-5 predicted verbs accuracies of both models marginally fluctuate.

The effect on the location of LayerNorm in GSRTR is shown at the third row of each set in Table~\ref{table:abla}.
GSRTR w/ Post-LN denotes that LayerNorm is placed between skip connections, leading to Post-Layer Normalization (Post-LN)~\cite{xiong2020layer} transformer architecture.
In all evaluation metrics of each set, the accuracies of GSRTR w/ Post-LN decrease.

\subsection{More Qualitative Results of Our Model}
\label{supp:pred}
\renewcommand\thefigure{A\arabic{figure}}

\begin{figure}[h!]
    \centering
        \includegraphics[width=0.99\textwidth]{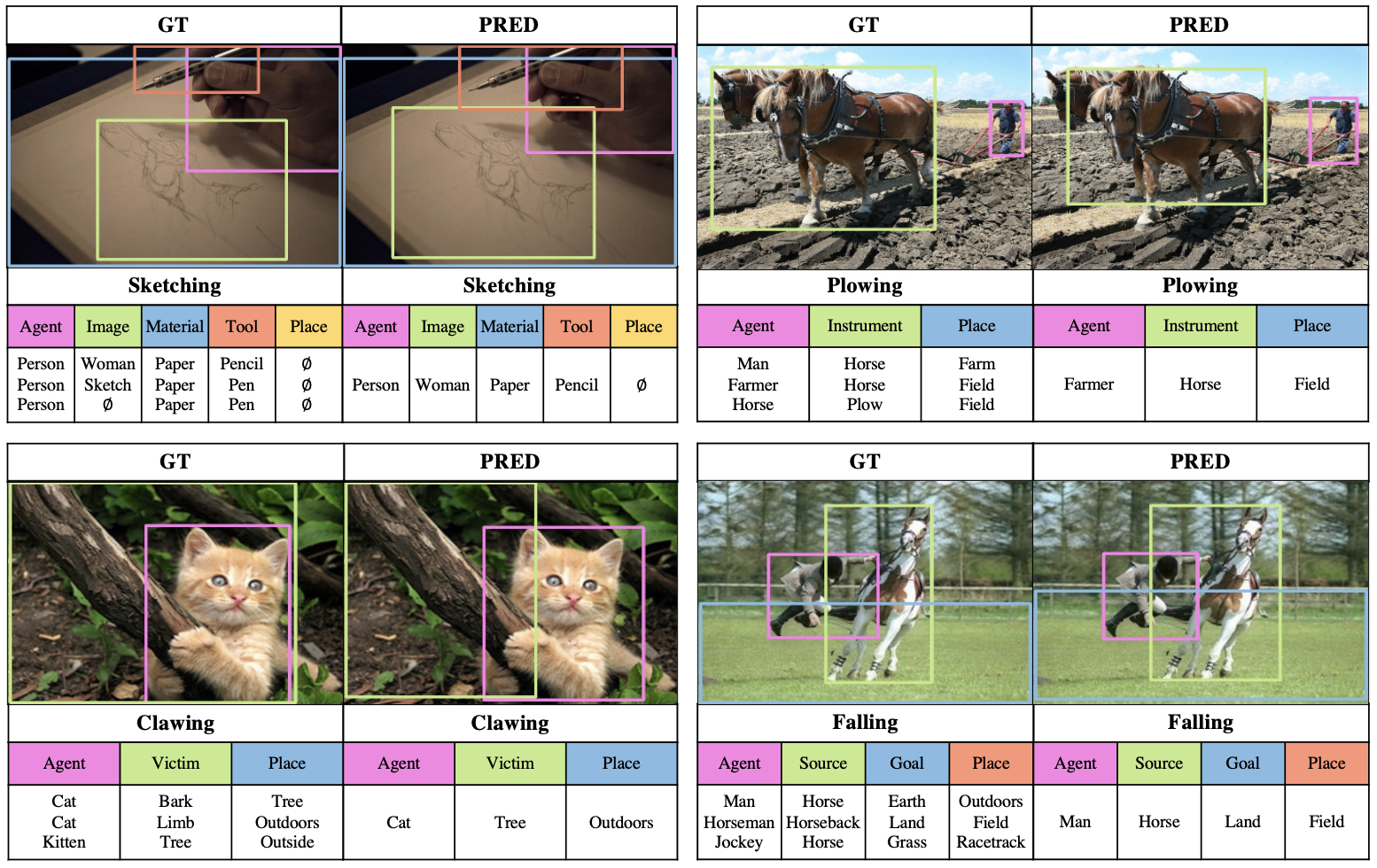}
\caption{
Correct grounded noun predictions of GSRTR in top-1 predicted verb setting on the \emph{test} set.
For each semantic role, three annotators record noun annotations.}
\label{fig:correct}
\end{figure}

\begin{figure}[h!]
    \centering
        \includegraphics[width=0.99\textwidth]{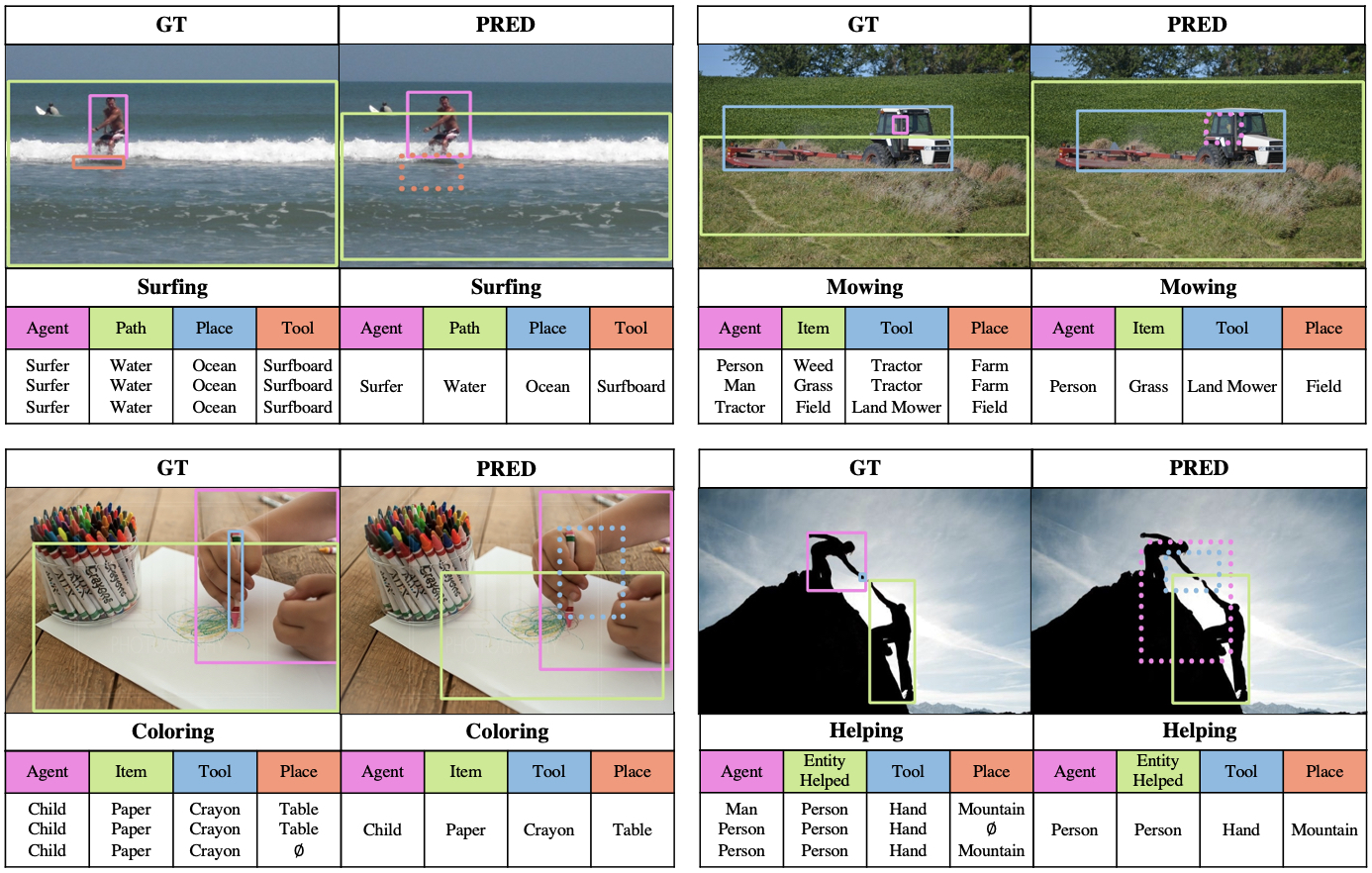}
\caption{
Incorrect box predictions of GSRTR in top-1 predicted verb setting on the \emph{test} set.
The dashed box denotes incorrect box prediction.
}
\label{fig:wrong_bbox}
\end{figure}

\begin{figure}[h!]
    \centering
        \includegraphics[width=0.965\textwidth]{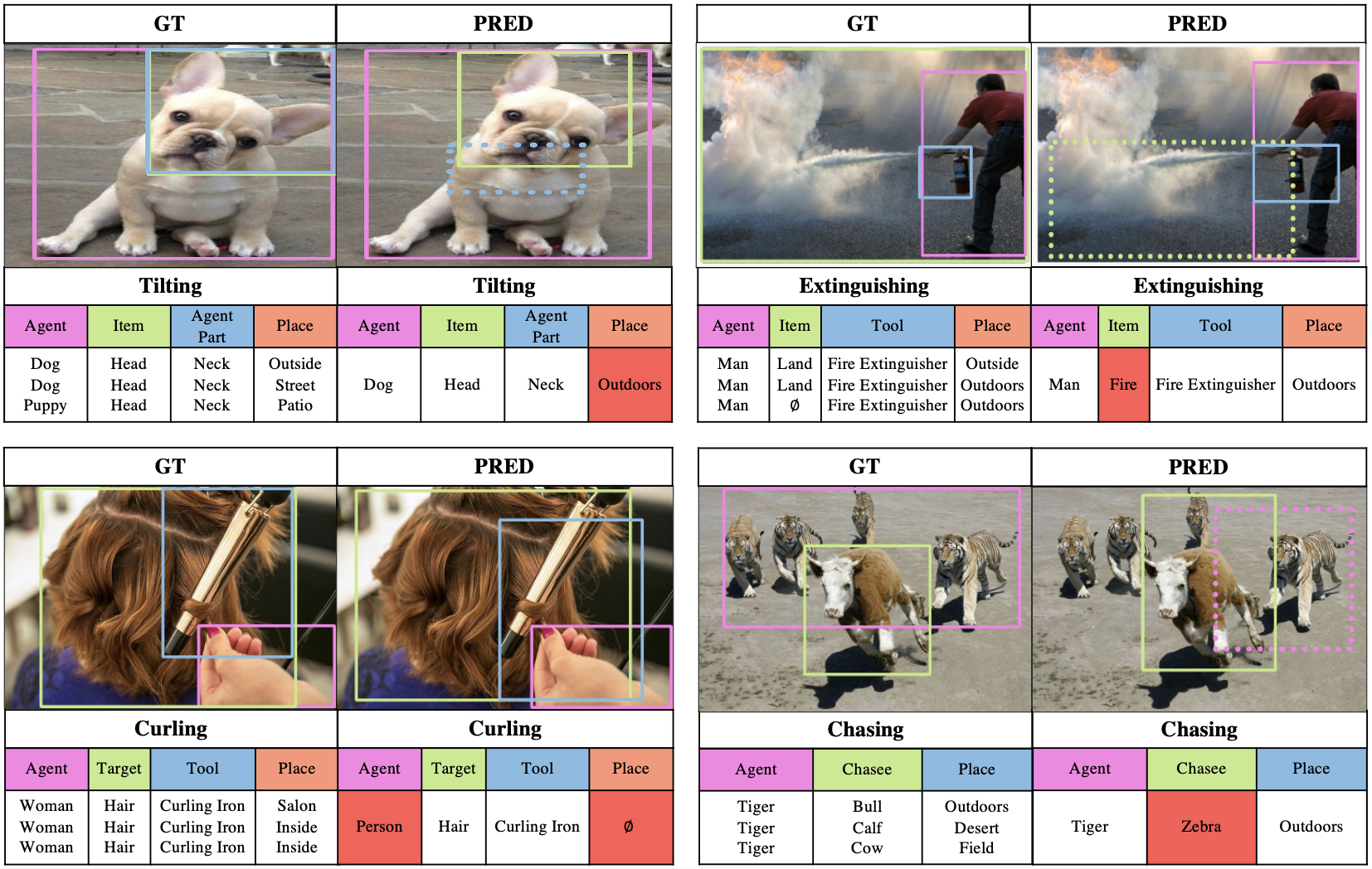}
\caption{
Incorrect noun predictions of GSRTR in top-1 predicted verb setting on the \emph{test} set.
The incorrect noun predictions are highlighted in red color.
The dashed box denotes incorrect box prediction.
}
\label{fig:wrong_noun}
\end{figure}

In top-1 predicted verb setting on the \emph{test} split of the SWiG dataset, the prediction results of GSRTR are shown in Figure~\ref{fig:correct}, Figure~\ref{fig:wrong_bbox} and Figure~\ref{fig:wrong_noun}.
The SWiG dataset has three noun annotations for each semantic role.
The noun prediction is considered correct if the predicted noun matches one of the three noun annotations.
The box prediction is considered correct if the model correctly predicts box existence and the predicted box has an Intersection-over-Union (IoU) value of at least 0.5 with the ground-truth box.
Note that the grounded noun prediction is considered correct if the predicted noun and predicted box are correct. 

Figure~\ref{fig:correct} shows the correct grounded noun prediction results.
Figure~\ref{fig:wrong_bbox} shows the failure cases of box prediction. 
There are incorrect box predictions when bounding boxes 
have extreme aspect ratios (\eg, the boxes of the role \textit{Tool} in the \textit{Surfing} and the \textit{Coloring} image), or small scales (\eg, the box of the role \textit{Agent} in the \textit{Mowing} image and the box of the role \textit{Tool} in the \textit{Helping} image).
Figure~\ref{fig:wrong_noun} shows the failure cases of noun prediction, including incorrect box predictions.
Even in the failure cases, there are the cases where GSRTR reasonably predicts nouns.
For example, 
in the \textit{Tilting} image, GSRTR predicts that the noun of the role \textit{Place} is \textit{Outdoors}, which is similar to the first annotation \textit{Outside}. 
In the \textit{Curling} image, GSRTR predicts that the nouns of the role \textit{Agent} and \textit{Place} are \textit{Person} and $\emptyset$, which are enough to describe the given image.
There is also the case where GSRTR inappropriately predicts nouns.
In the \textit{Chasing} image, GSRTR predicts that the noun of the role \textit{Chasee} is \textit{Zebra}, whereas the three noun annotations are \textit{Bull}, \textit{Calf}, and \textit{Cow}.

\clearpage

\renewcommand\thefigure{A\arabic{figure}}

\begin{figure}[h!]
    \centering
        \includegraphics[width=0.96\textwidth]{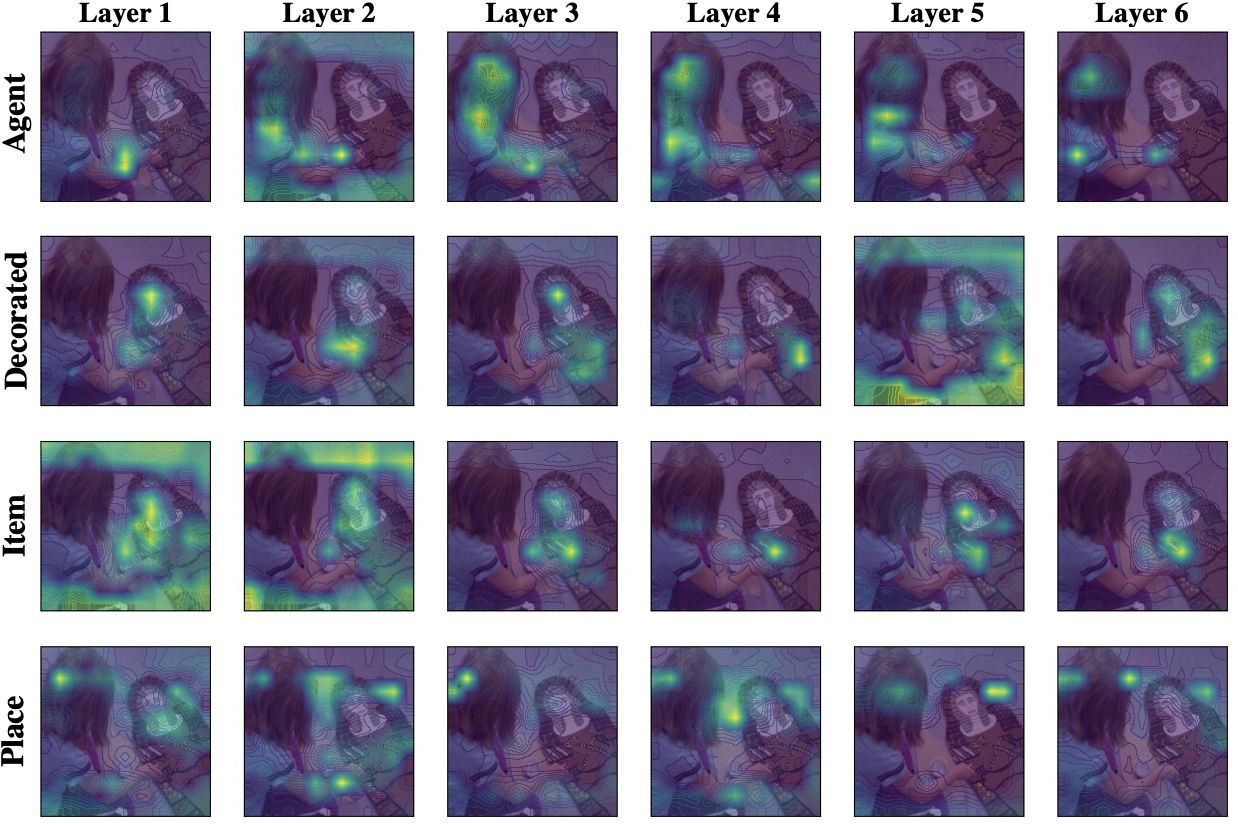}
\caption{
    Role Attention Map on Image Features for a \textit{Decorating} image from the MHA block in each decoder layer.
}
\label{fig:decorating}
\end{figure}

\subsection{Qualitative Analysis on Attention}
\label{supp:viz}

\begin{figure}[h!]
    \centering
        \includegraphics[width=0.97\textwidth]{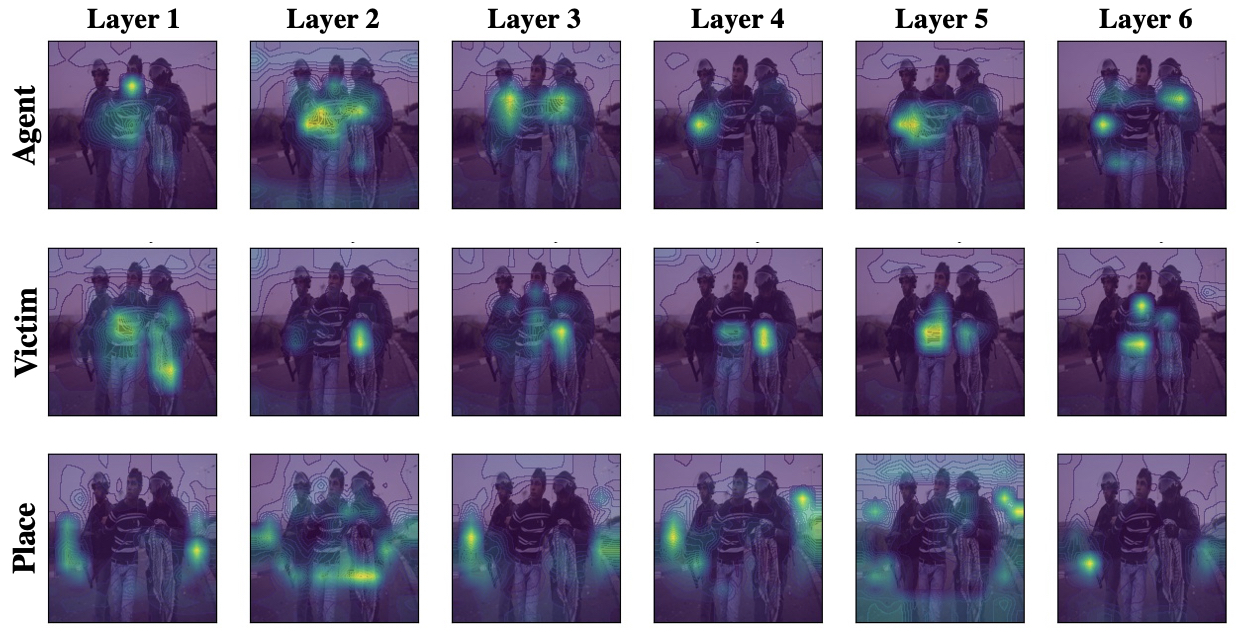}
\caption{
    Role Attention Map on Image Features for a \textit{Apprehending} image from the MHA block in each decoder layer.
}
\label{fig:apprehending}
\end{figure}

\begin{figure}[h!]
    \centering
        \includegraphics[width=0.97\textwidth]{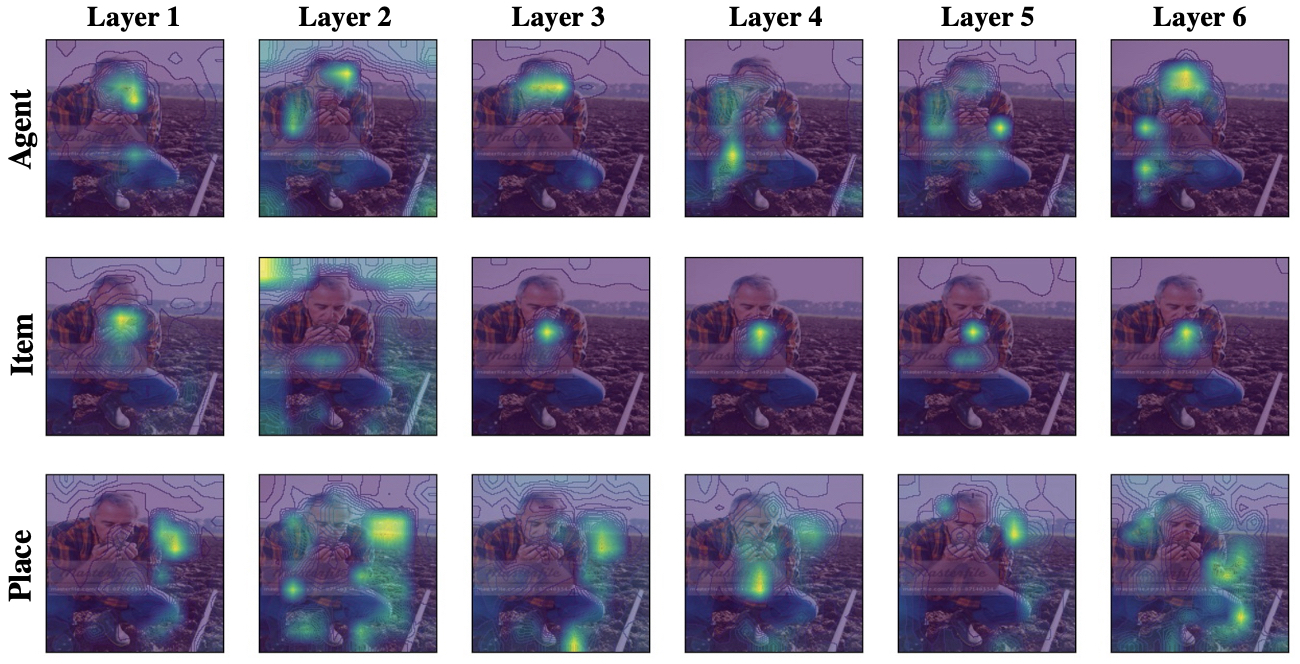}
\caption{
    Role Attention Map on Image Features for a \textit{Smelling} image from the MHA block in each decoder layer.
}
\label{fig:smelling}
\end{figure}

\noindent
\textbf{Role Attention Map on Image Features:}
In Figure~\ref{fig:decorating}, Figure~\ref{fig:apprehending} and Figure~\ref{fig:smelling}, each column shows the difference of attention maps among roles.
Each row shows the transition of attention maps through the decoder layers.
In Figure~\ref{fig:decorating}, the role \textit{Decorated} focuses on the decorated stuff and the role \textit{Item} focuses on the decoration item.
Figure~\ref{fig:apprehending} shows that GSRTR can understand the given image and distinguish between the role \textit{Agent} and the role \textit{Victim}.
Figure~\ref{fig:apprehending} and Figure~\ref{fig:smelling} show that GSRTR can figure out the background for the role \textit{Place} in the given image.

\begin{figure}[h!]
    \centering
        \includegraphics[width=0.99\textwidth]{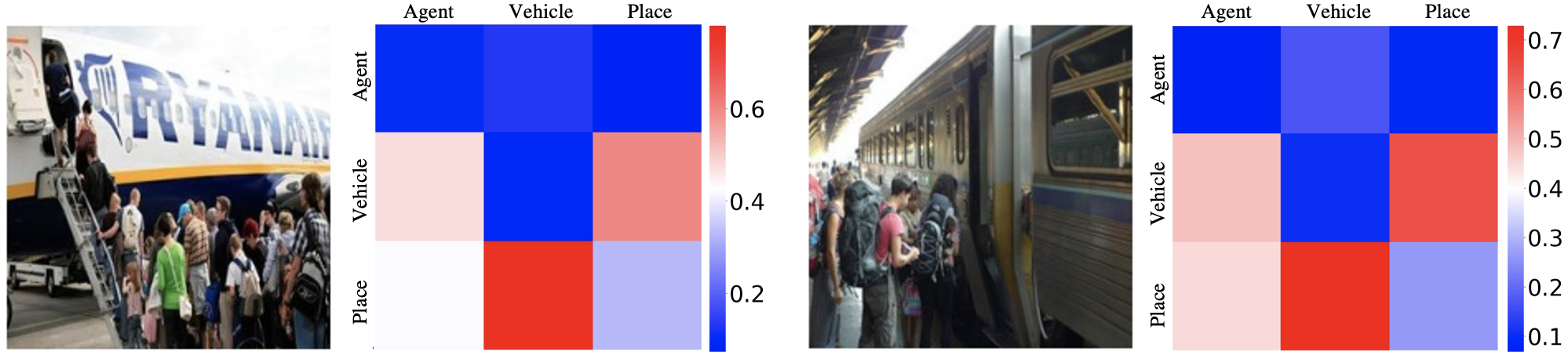}
\caption{
    Visualization on Role Relations for two \textit{Boarding} images from the MHSA block in the last decoder layer.
    Attention scores are represented as column-wise sum to 1.
}
\label{fig:boarding}
\end{figure}

\begin{figure}[h!]
    \centering
        \includegraphics[width=0.99\textwidth]{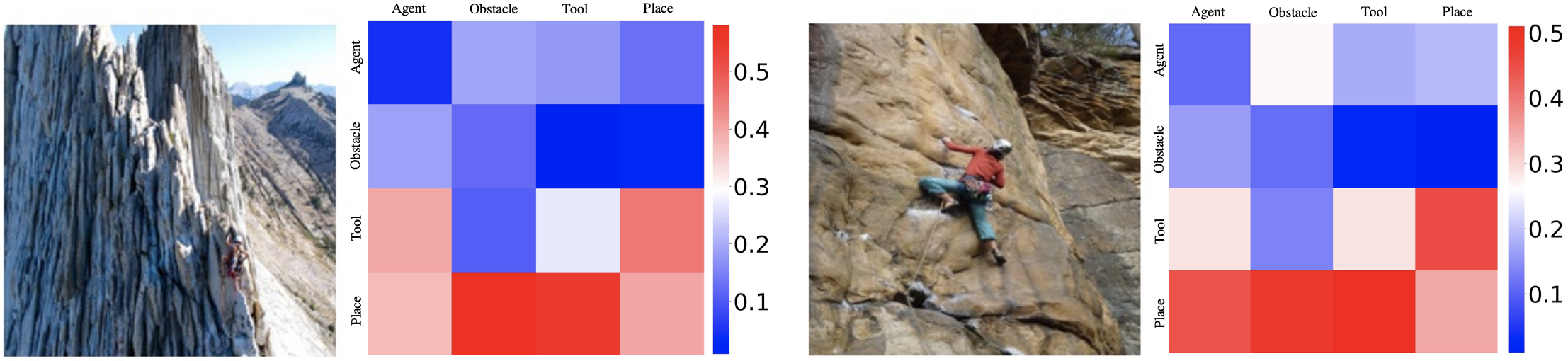}
\caption{
    Visualization on Role Relations for two \textit{Climbing} images from the MHSA block in the last decoder layer.
    Attention scores are represented as column-wise sum to 1.
}
\label{fig:climbing}
\end{figure}

\noindent
\textbf{Visualization of Role Relations:} GSRTR captures the relations among roles in the similar way if the situations of the given images are similar. In Figure~\ref{fig:boarding}, the role \textit{Vehicle} focuses on the role \textit{Place}, \emph{i.e.}, the runway (\textit{Place}) and the railway station (\textit{Place}) are highly related to the airplane (\textit{Vehicle}) and the train (\textit{Vehicle}) given the verb \textit{Boarding}, respectively. In Figure~\ref{fig:climbing}, the role \textit{Obstacle} and the role \textit{Tool} focus on the role \textit{Place}, \emph{i.e.}, the cliff (\textit{Place}) is highly related to the rock (\textit{Obstacle}) and the rope (\textit{Tool}) given the verb \textit{Climbing}.


\begin{figure}[!htb]
    \centering
        \includegraphics[width=0.99\textwidth]{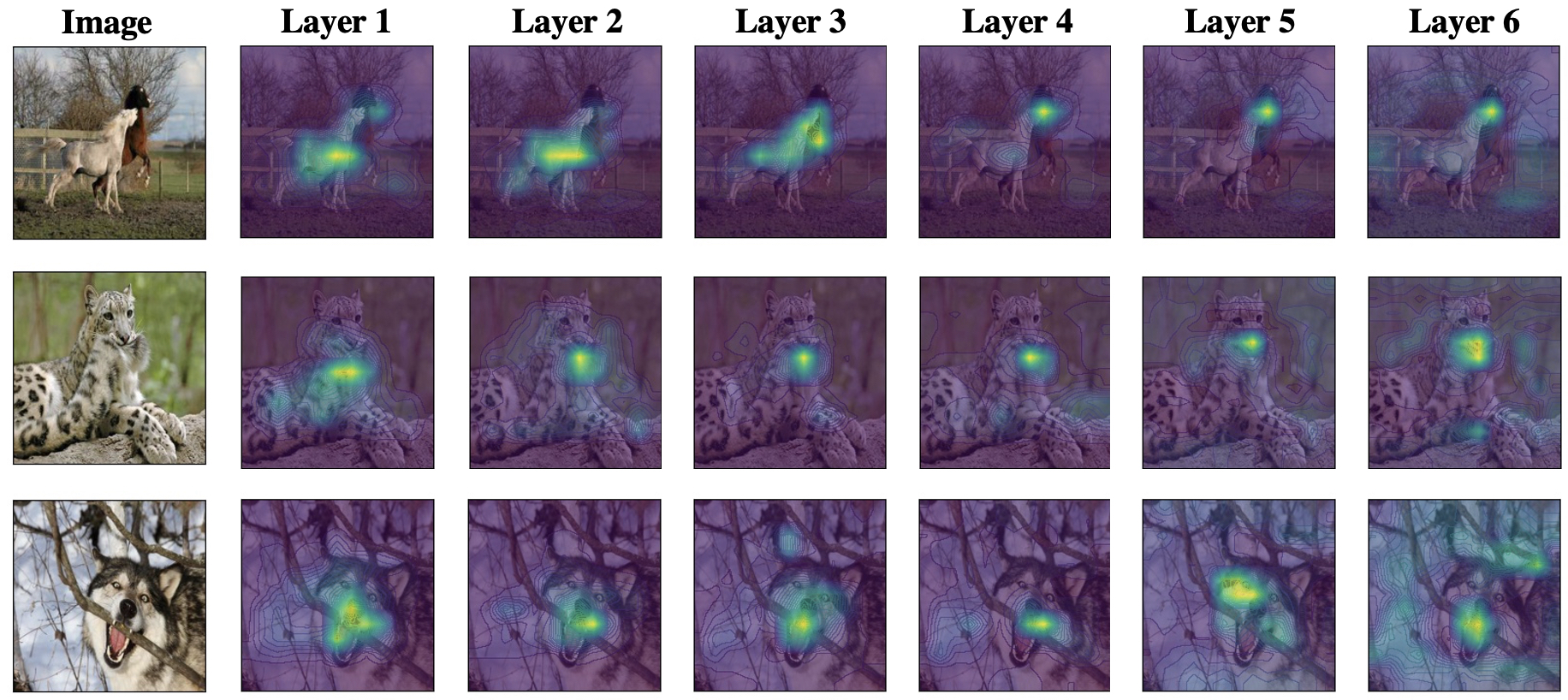}
\caption{
    Verb Token Attention Map on Image Features for three \textit{Biting} images.
    Each row consists of an image and attention maps from the MHSA block in each encoder layer.
}
\label{fig:biting}
\end{figure}

\begin{figure}[!htb]
    \centering
        \includegraphics[width=0.99\textwidth]{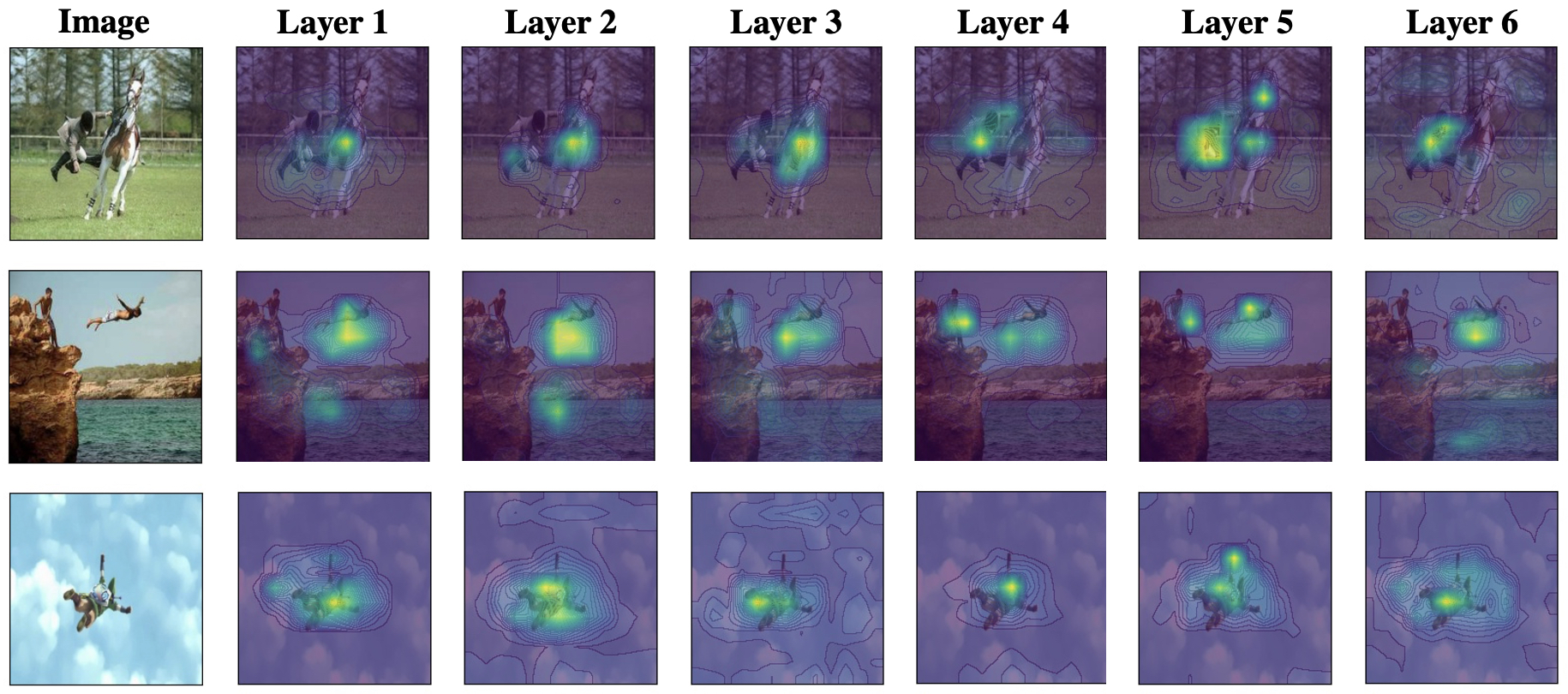}
\caption{
    Verb Token Attention Map on Image Features for three \textit{Falling} images.
    Each row consists of an image and attention maps from the MHSA block in each encoder layer.
}
\label{fig:falling}
\end{figure}

\newpage
\noindent
\textbf{Verb Token Attention Map on Image Features:} GSRTR can capture the key feature to infer the salient action. Figure~\ref{fig:biting} and Figure~\ref{fig:falling} show that GSRTR focuses on the bitten part and the falling agent, respectively. The rightmost column shows that the semantic regions where the verb token focuses on are similar for the same verb. Each row shows the transition of attention maps through the encoder layers.

\end{document}